\documentclass{article}

\usepackage{PRIMEarxiv}

\usepackage[utf8]{inputenc} 
\usepackage[T1]{fontenc}    
\usepackage{hyperref}       
\usepackage{url}            
\usepackage{booktabs}       
\usepackage{amsfonts}       
\usepackage{nicefrac}       
\usepackage{microtype}      
\usepackage{lipsum}
\usepackage{fancyhdr}       
\usepackage{graphicx}       
\usepackage{todonotes}
\usepackage{amsmath}
\usepackage{float}
\usepackage{algpseudocode}
\usepackage{algorithm}
\usepackage{multirow}
\graphicspath{{media/}}     
\usepackage{amssymb}
\usepackage{pifont}
\usepackage{enumerate}
\usepackage[inline]{enumitem}
\usepackage{natbib}
\usepackage{longtable}
\usepackage{caption}
\newcommand{\ph}[1]{\textcolor{black}{#1}}

\newcommand{\paco}[1]{\textcolor{black}{#1}}

\newcommand{\major}[1]{\textcolor{black}{#1}}

\newcommand{\rewrite}[1]{\textcolor{black}{#1}}
\newcommand{\minor}[1]{\textcolor{black}{#1}}

\newenvironment{minorenv}{\color{black}}{}                

\newenvironment{colortabular}[1]
  {\begingroup\color{#1}\ignorespaces}
  {\endgroup\ignorespacesafterend}

\title{From Privacy to Trust in the Agentic Era: A Taxonomy of Challenges in Trustworthy Federated Learning Through the Lens of Trust Report 2.0
}

\author{
  Nuria Rodríguez-Barroso, Mario García-Márquez \\
  Department of Computer Science and Artificial Intelligence,\\ Andalusian Research Institute in Data Science \\
  and Computational Intelligence (DaSCI) \\
  University of Granada \\
  Granada\\
  \texttt{\{rbnuria, mariogmarq\}@ugr.es} \\
   \And
  M.V. Luzón \\
    Department of Software Engineering,\\ 
    Andalusian Research Institute in Data Science \\
  and Computational Intelligence (DaSCI) \\
  University of Granada \\
  Granada\\
  \texttt{luzon@ugr.es} \\
  \And
  Francisco Herrera \\
  Department of Computer Science and Artificial Intelligence,\\ Andalusian Research Institute in Data Science \\
  and Computational Intelligence (DaSCI) \\
  University of Granada \\
  Granada\\
  \texttt{herrera@decsai.ugr.es} \\
}

\begin{document}
\maketitle

\begin{abstract}
\ph{Federated Learning (FL) enables privacy-preserving collaborative learning, yet deployments increasingly show that privacy guarantees alone do not sustain trust in high-risk settings. As FL systems move toward agentic AI, large language model–enabled, and dynamically adaptive architectures, trustworthiness becomes a system-level problem shaped by autonomous decision-making, non-stationary environments, and multi-stakeholder governance. We argue for Trustworthy FL (TFL), treating trust as a continuously maintained operating condition rather than a static model property.}

\ph{Through the lens of Trust Report~2.0, we propose a requirement-driven taxonomy of challenges grounded in TAI and explicitly extended to account for control-plane decisions, agency, and system dynamics across the federated lifecycle. Building on this diagnosis, we introduce a coordination blueprint that structures cross-requirement trade-offs, decision justification, and governance alignment in TFL systems. To operationalize assurance, Trust Report~2.0 is instantiated as a lightweight, privacy-preserving artifact that surfaces decision-centric trust evidence without centralizing raw data. We illustrate applicability via healthcare as a stress-test domain, focusing on oncology FL under regulatory pressure and clinical risk.}
\end{abstract}

\section{Introduction: From Privacy Guarantees to Trust Evidence}\label{sec1:intro}

The increasing deployment of Artificial Intelligence (AI) systems in sensitive domains \cite{islam2022systematic} such as healthcare, finance, and law enforcement, has intensified the need for frameworks that guarantee ethical, legal, and technical alignment with societal values. In response, the notion of trustworthy AI \paco{(TAI)} \cite{li2023Trustworthy} has emerged as a foundational principle for the development and deployment of AI. \paco{This concept has been articulated by prominent bodies such as the European Commission, as part of the Ethical Guidelines~\cite{eu2019ethics}, and the National Institute of Standards and Technology~\cite{nist2023artificial}. For the purposes of this work, we will mainly refer to the characterization provided by the former}. As described in the European Commission Ethics Guidelines, \paco{TAI} is characterized by the adherence to seven key requirements \cite{diaz2023connecting}: (1) human agency and oversight, (2) technical robustness and safety, (3) privacy and data governance, (4) transparency, (5) diversity and fairness, (6) societal and environmental well-being, and (7) accountability.

\ph{Within this landscape, FL  \cite{kairouz2021advances} has emerged as a foundational paradigm for collaborative model training across distributed and often sensitive data silos, enabling learning without explicit data centralization. Initially motivated by privacy preservation and regulatory constraints, FL has been widely adopted in domains such as mobile systems, healthcare, finance, and industrial Internet of Things (IoT). Over the past decade, a rich body of work has addressed core challenges related to communication efficiency, statistical heterogeneity, privacy leakage, robustness to adversarial behavior, and system scalability. As a result, privacy-preserving mechanisms such as secure aggregation and differential privacy are now considered baseline components of many FL deployments \cite{li2021survey}.}

\ph{FL was originally motivated by the need to enable collaborative model training under strict privacy and data governance constraints, particularly in domains such as healthcare or finance. However, experience from real-world deployments has shown that privacy preservation alone is insufficient to sustain trust in complex, high-risk settings.~\cite{kairouz2021advances,kanauzia2026comprehensive}.}

\ph{This claim exposes a critical gap in the existing literature. While numerous surveys and overviews have catalogued FL challenges, often focusing on optimization, communication, or privacy, most treat trust implicitly or narrowly, typically equating it with confidentiality or robustness against specific attacks. Even recent works on TFL tend to emphasize isolated dimensions such as security or fairness, without providing a unifying framework that accounts for agency, autonomy, governance, and continuous assurance under real-world constraints.}

\ph{Meanwhile, emerging research on agentic AI and its adoption in high-stakes domains (e.g., healthcare~\cite{hanna2026infectious}) increasingly emphasizes the need for stronger accountability, human oversight, and evidence-based governance mechanisms. Recent work distinguishes between \emph{AI agents} (typically single-entity, tool-augmented systems) and \emph{agentic AI}, which denotes coordinated, multi-agent systems characterized by persistent autonomy, dynamic task decomposition, shared memory, and emergent behavior \cite{SAPKOTA2026103599}. This distinction is especially relevant for FL, where decision-making authority is inherently distributed across participants, infrastructures, and governance layers. When such agentic capabilities are embedded into FL settings, traditional assumptions that trustworthiness can be ensured primarily through privacy guarantees or pointwise model properties become increasingly untenable. Instead, trust must account for system-level autonomy, inter-agent coordination, and lifecycle governance across both learning and control planes.}

\ph{In parallel, recent advances in agentic AI and large language models (LLMs) are fundamentally reshaping the capabilities and risk profile of FL deployments. Agentic AI systems are no longer passive optimizers that only update model parameters; instead, they can autonomously plan, select actions, invoke tools, adapt objectives, and coordinate with other agents to pursue complex goals~\cite{acharya2025agentic,jiang2025large}. When such capabilities are embedded into federated settings, where decisions are distributed across organizations, jurisdictions, and infrastructures, the traditional assumption that trustworthiness can be ensured primarily through privacy guarantees becomes increasingly untenable.}

\ph{At the same time, the integration of foundation models and LLMs into federated and distributed learning pipelines introduces new trust challenges. LLMs amplify concerns around memorization, semantic leakage, alignment with human values, and unpredictable emergent behaviors, particularly when adapted continuously across heterogeneous clients \cite{wei2025federated,amini2025distributed}. In parallel, many real-world FL deployments operate in dynamic and non-stationary environments, where data distributions, client populations, and system conditions evolve over time. In such settings, trustworthiness is no longer a static property of a trained model but a temporal and lifecycle-dependent condition that can degrade due to concept drift, client churn, or autonomous system decisions \cite{polato2026learning}. }

\ph{Against this backdrop, this paper argues that trustworthiness in FL must be reconceptualized for the agentic AI era. Privacy remains a necessary precondition for collaboration, but it is no longer sufficient to guarantee that federated systems behave safely, fairly, transparently, and accountably over time. Instead, trust must be understood as a system-level, continuously evidenced property that spans both the learning plane (model training and aggregation) and the control plane (decisions about participation, objectives, evaluation, deployment, and adaptation).}

\ph{Accordingly, this paper is guided by  four fundamental questions:}

\begin{itemize}
 \item [\textbf{Q4}] \ph{\textit{Structured Diagnosis toward TFL.} What are the fundamental challenges that prevent FL systems from being trustworthy in the AI agentic AI era, and how can these challenges be systematically organized under the principles of TAI?}
 
 \item [\textbf{Q2}] \ph{\textit{Technical Operationalization.} How can trustworthiness in FL be operationalized and sustained over time-beyond pointwise guarantees through system design, agency management, and auditable lifecycle evidence?}
 \item [\textbf{Q3}] \ph{\textit{Paradigm Shift,} How does the notion of trustworthiness in FL change when moving from static, privacy-centric systems to agentic AI, LLM-enabled systems operating in dynamic environments?}
   \item[\textbf{Q4}] \ph{\textit{Governance and Accountability.} What governance mechanisms, accountability structures, and technical evidence are required to assign responsibility and ensure regulatory compliance in agentic, multi-stakeholder FL systems? }

\end{itemize}

\ph{To answer these questions, we make four main contributions.}

\begin{itemize}
 \item \ph{First, we present a requirement-driven challenges taxonomy grounded in TAI principles, explicitly extended to account for agentic behavior, LLM integration, and dynamic environments. This taxonomy highlights not only individual challenges but also the tensions and trade-offs that arise when multiple trust dimensions interact under decentralization and autonomy (Q1).}

 \item \ph{Second, we propose a coordination blueprint for TFL, that introduces an explicit separation between learning and control planes, incorporates agency-aware design, and aligns technical mechanisms with governance and oversight requirements (Q2).}

\item \ph{Third, we provide a conceptual re-framing of trustworthiness in FL for the agentic AI era, showing how the transition from static, privacy-centric training pipelines to agentic AI, LLM-enabled systems operating in dynamic environments fundamentally alters the trust problem. In particular, we argue that trust can no longer be treated as a static property of trained models or isolated guarantees, but must instead be understood as a system-level, lifecycle-dependent condition shaped by autonomous decision-making, evolving objectives, and environmental non-stationarity. This paradigm shift motivates an explicit agency-aware perspective, including the separation between the learning plane and the control plane, a decomposition of agency across decision loci in the federated lifecycle, the characterization of different levels of autonomy, and the identification of agency-specific threat models. Together, these elements establish the foundation for continuous trust evidence and governance throughout the FL lifecycle (Q3).}

 \item \ph{Fourth, we introduce a lightweight yet extensible \textit{Trust Report} as an operational assurance artifact, designed to provide continuous, auditable evidence of trustworthiness across the FL lifecycle. Therefore, we move from mapping challenges to reframing Trust itself in federated settings. Building on recent work on Trust in AI \cite{HENRIQUE2024100043}, we treat Trust not as a static label but as a context-dependent operating condition that is co-produced by technical safeguards and governance roles in TFL.
  We demonstrate the applicability of this blueprint through a healthcare case study, focusing on cross-institutional settings where regulatory constraints, safety requirements, and environmental dynamics place particularly stringent demands on trust (Q4).}

\end{itemize}

\ph{By moving from privacy to trust as the central design objective, this work provides a unified conceptual and practical foundation for FL systems operating in the agentic AI era. The proposed taxonomy and blueprint aim to support researchers, system designers, and regulators in building FL deployments that remain trustworthy not only at training time, but throughout their operational lifetime.}

\ph{TFL is achieved through \emph{decision-centric governance} that links TAI requirements to FL-native controls and auditable evidence, across both learning and control planes, over the system lifecycle.}

\ph{We provide a concise synthesis of prior survey efforts on open challenges in FL, with the explicit goal of identifying recurring gaps and structural limitations that motivate our perspective. Existing surveys offer valuable insights into privacy, robustness, or fairness in isolation, yet largely treat trustworthiness as a static, model-centric property and seldom account for agency, control-plane decisions, or system-level dynamics in evolving FL architectures. To preserve focus on these limitations and their implications, extended background discussions and complementary technical details are consolidated in Appendix A, where they remain accessible without interrupting the progression toward a requirement-driven, agency-aware, and decision-centric formulation of trustworthiness. }

\ph{The remainder of the paper is organized as follows. 
In Section~\ref{sec2:preliminaries}, we introduce the essential FL background concepts required for the rest of the paper. 
In Section~\ref{sec3:challenges}, we present the seven TAI requirements and develop a taxonomy of challenges for aligning FL systems with these principles; for each requirement, we further identify open challenges and analyze the relevant literature. 
Section~\ref{sec:Agentic_Dynamic} formalizes how trustworthiness is reshaped in agentic AI and dynamic FL settings, highlighting the implications of autonomy, evolving objectives, and environmental non-stationarity. 
In Section~\ref{Sec:TAI-Blueprint}, we propose a coordination blueprint that surfaces cross-requirement interactions in TFL and introduces concrete coordination mechanisms with particular emphasis on auditability and governance, through the lens of Trust Report 2.0. 
Section~\ref{sec:TAI-Health} presents a oncology scenario, providing a practitioner-oriented mapping between the proposed TAI requirements and concrete Trust Report signals and governance actions. 
In Section~\ref{Sec:TAI-Privacty-to-Trust}, we reframe trust in FL as a contextual and continuously maintained operating condition, tightly coupled with governance processes. 
Finally, Section~\ref{sec:final} concludes the paper and outlines key takeaways and future research directions. In Appendix A we provide a concise synthesis of prior survey efforts on open challenges, highlighting the limitation that motivate our perspective. }

\section{Background on Federated Learning}
\label{sec2:preliminaries}

\ph{\minor{In this section, we introduce the FL setting including its motivation, training protocol and basic formalism.}}

The increasing data volume and diversity requirements have led to challenges concerning data privacy and the processing of large datasets. FL emerges as a solution to address these issues, particularly focusing on privacy, communication, and data accessibility.

\paragraph{\paco{Why?}}
\begin{itemize}
    \item \textit{Data Privacy}: In traditional centralized ML, user data is aggregated and stored on central servers, increasing the risk of privacy violations \cite{attacks_CV}. This concern is especially pronounced in sectors such as healthcare and finance, where data sensitivity is paramount \cite{abouelmehdi2017big}. Furthermore, stringent data protection regulations, such as \paco{European} General Data Protection Regulation (GDPR) \cite{gdpr2}, require the development of AI methodologies that preserve privacy.
    \item \textit{Communication Costs and Latency} \cite{bib:mcmahan16communicationefficient}: Centralized ML often involves transmitting raw data to central servers for processing and model training, which can be resource-intensive and time-consuming, especially with large datasets. The proliferation of IoT devices has further exacerbated this challenge, as the continuous flow of data from diverse sources demands efficient storage and preprocessing solutions.
    \item \textit{Limitations in Data Access} \cite{abouelmehdi2017big}: Data are frequently distributed across various institutions or organizations, preventing seamless access or sharing due to legal, regulatory, or technical constraints. This fragmentation poses challenges for centralized ML approaches that rely on consolidated datasets for effective model training.
\end{itemize}

\paragraph{\paco{How?}}

In this context arises FL \cite{yang2019federated}, a distributed ML paradigm that enables the development of a global model without the need to exchange raw data among participants. This approach involves a network of clients, denoted as $\{C_1, C_2, \dots, C_n\}$, and operates primarily in two phases:

\begin{enumerate}
    \item \textit{Model Training Phase}: Each client trains a local model on its own data and shares only the model updates, not the raw data. These local models are then aggregated to form a global model, ensuring data privacy is maintained throughout the process.
    \item \textit{Inference Phase}: The aggregated global model is used to make predictions on new data instances.
\end{enumerate}

\begin{figure}[h!]
  \centering
  \includegraphics[width=0.6\textwidth]{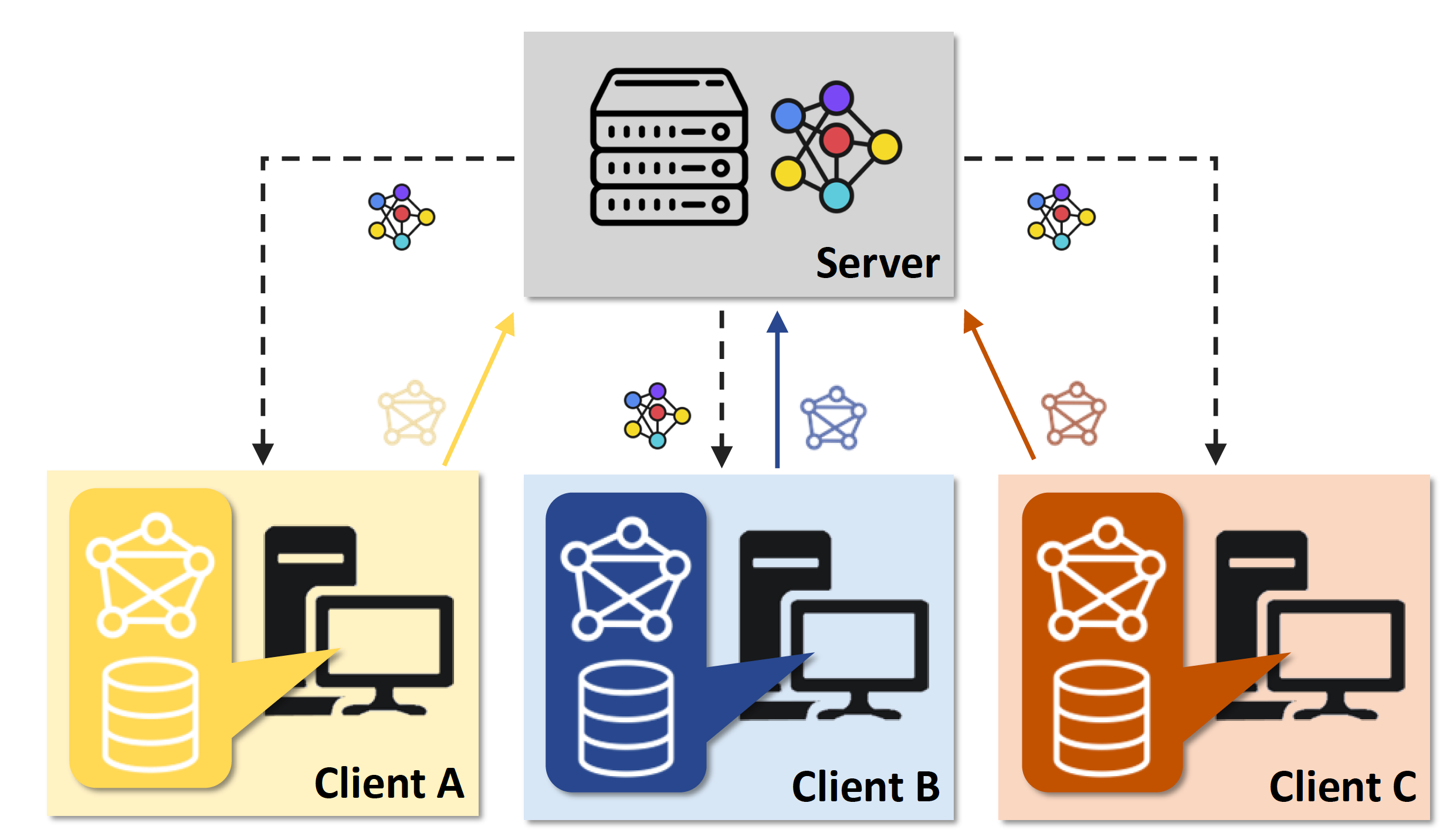}
  \caption{Representation of the round in FL. Figure inspired by \cite{luzon2024tutorial}.}
  \label{fig:FL}
\end{figure}

These processes can be executed synchronously or asynchronously, depending on the availability of data and the specific requirements of the model. It is important to note that beyond privacy preservation, establishing a fair value-distribution mechanism is crucial to equitably share the benefits derived from the collaboratively trained model. We provide a visual representation of this learning process in Figure~\ref{fig:FL}.

Formally, a FL scenario can be described as follows: Consider a set of clients or data owners $\{C_1, \dots, C_n\}$, each of whom has local training data $\{D_1, \dots, D_n\}$. Each client $C_i$ maintains a local learning model $L_i$ with parameters $\{L_1, \dots, L_n\}$. The objective of FL is to learn a global model $G$ leveraging distributed data between clients through an iterative process known as a \textit{learning round}. In each learning round $t$:

\begin{enumerate}
    \item Each client trains its local model on its respective local data $D^t_i$, updating its parameters from $L^{t}_i$ to $\hat{L}^t_i$.
    \item Global parameters $G^t$ are calculated by aggregating updated local parameters $\{\hat{L}^t_1, \dots, \hat{L}^t_n\}$ using a predefined federated aggregation operator $\Delta$:
    \begin{equation}
    \begin{split}
        G^t = \Delta(\hat{L}^t_1,\hat{L}^t_2, \dots, \hat{L}^t_n)\\
        L^{t+1}_i \leftarrow G^t, \quad \forall i \in \{1, \dots, n\}.
    \end{split}
    \label{eq_fl_aggregation}
    \end{equation}
\end{enumerate}

This iterative update continues until a specified stopping criterion is met, resulting in a global model $G$ that encapsulates the collective knowledge of all participants.

\ph{Additional background and complementary technical considerations  to position with respect the existing reviewers are provided in \ref{sec:related_work} to preserve focus on the system-level trust framework introduced in the following.}

\section{A Requirement-Driven Taxonomy of Trust-Critical Challenges in Federated Learning}\label{sec3:challenges}

Given the growing emphasis on ensuring that AI systems are ethically sound, legally compliant and technically robust, aligning FL with the requirements of {TAI is essential. These requirements provide a structured framework to evaluate whether FL systems can be considered reliable and responsible in real-world applications. Therefore, in the following sections, we present the main challenges of aligning FL with the requirements of TAI, organizing them according to the key requirements of TAI \cite{diaz2023connecting}. This approach allows us to highlight where challenges arise and where further research is needed to ensure that FL contributes effectively to the development of human-centered TAI. The main challenges taxonomy to align FL with TAI is reflected in Figure~\ref{fig:example}.

\begin{figure}[h!]
  \centering
  \includegraphics[width=0.8\textwidth]{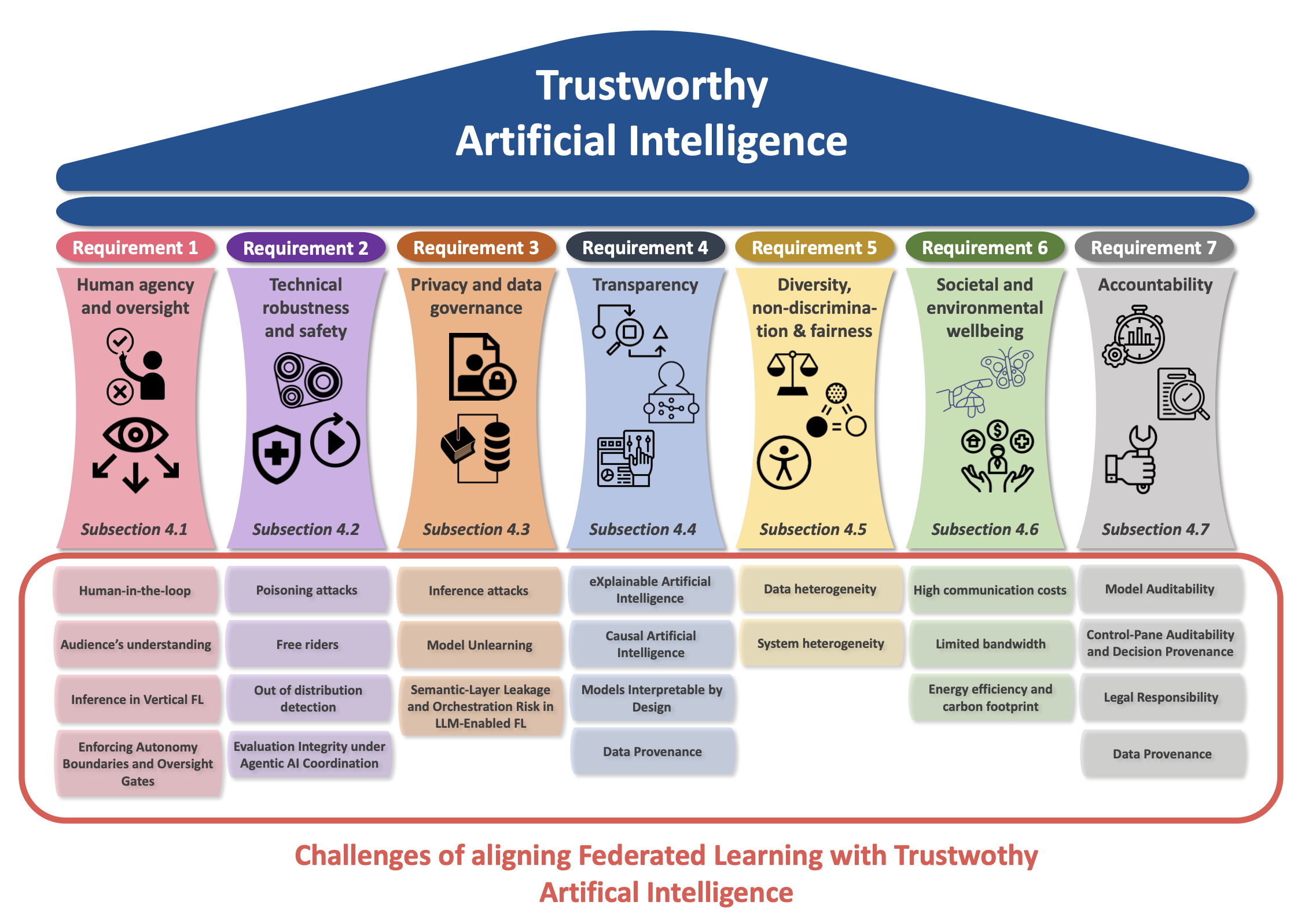}
  \caption{Taxonomy of challenges to  align FL with \paco{TAI}. Inspired in \cite{diaz2023connecting}, \paco{under} each \paco{TAI} requirement, we list the specific challenges involved to meet the requirement within an FL paradigm.}
  \label{fig:example}
\end{figure}

The organization of this section is as follows. First, we introduce each requirement and explain how FL naturally satisfies it, emphasizing the facets that are fulfilled by definition. We then identify the challenges that are intrinsic to FL when trying to meet that requirement. Finally, for each requirement, we highlight the following three categories:

\begin{itemize}
    \renewcommand{\labelitemi}{\ding{72}}

    \item \textit{Done}: issues that have already been addressed in the literature. \major{This can be seen as the "classical work" within a given challenge. These works typically establish foundational principles, propose baseline methodologies, and set reference points for comparison. They provide the groundwork upon which current and future research builds, often being cited as key contributions in the field.}

    \item \textit{Trends}: the main research directions currently under active investigation. \major{These encompass emerging approaches, techniques, and methodologies that are gaining momentum within the community. They often reflect the latest technological advancements or newly discovered insights that reshape existing paradigms. Identifying these trends helps contextualize where the field is heading and what areas are attracting the most attention.}

    \item \textit{To do}: technical aspects that remain unresolved or for which existing solutions are still incomplete. \major{This category is mainly formed by novel work which is mainly conceptual or future work identified by current trends. These topics represent open challenges that require innovative thinking, deeper theoretical understanding, or new empirical validation. They highlight potential research gaps and opportunities for future exploration.}
\end{itemize}

Some challenges do not contain all three categories. This is because certain lines of work have already been solved, others are only now beginning to be explored, and still others have yet to be addressed satisfactorily. \minor{We close each challenge with concise open research questions to explicitly surface the most salient remaining research gap.}

\major{The included references aim to provide a comprehensive overview for each challenge, thereby representing the global research landscape in FL through the lens of TAI and the proposed solutions. The categorization of these references into the aforementioned categories was predicated on both their publication date and the novelty of their contributions at the time of publication. Nevertheless, due to the disparity in research interest across the various challenges, some have received more extensive exploration than others. Consequently, publication dates for the first work in each challenge vary and thus the criteria for specific novelty and relevant publication dates become challenge-dependent and somewhat nuanced.}

\major{Finally, a short summary as Figure \ref{fig:key_challenges} and Table \ref{tab:key_gaps_short} included in subsection \ref{sec3:summary_table}. It shows graphically the remaining work and key gap together with the challenges taxonomy and blueprint in the table. It provides a global view of the section discussion. }

\subsection{Requirement 1: Human agency and oversight}

The first requirement of \paco{TAI}, human agency and oversight \cite{diaz2023connecting}, emphasizes that AI systems must empower human decision-making rather than undermine it. Effective oversight mechanisms are crucial to ensure that humans remain meaningfully involved throughout the AI lifecycle. In the context of FL, meeting this requirement involves ensuring that end-users and system operators can understand, guide, and override the system’s behavior when necessary, particularly in high stakes or safety critical applications.

\paragraph{\textbf{Challenge 1.1: Human-in-the-loop}} The integration of Human-in-the-Loop (HITL) principles into machine learning lifecycles, encompassing training, tuning, evaluation, and inference, has gained prominence due to the enhanced trustworthiness and accountability it offers by empowering human oversight in critical decision-making processes~\cite{wu2022survey}. However, the inherent distributed nature of FL introduces significant complexities for a straightforward implementation of HITL. For example, FL training typically involves numerous distributed devices, each potentially requiring human intervention~\cite{ng2021federated}, which presents substantial scalability challenges in terms of client management. Furthermore, the incorporation of human input into such systems can introduce new attack possibilities, potentially enabling the exploitation of human feedback loops or the injection of misleading annotations~\cite{hirai2023federated}.

\begin{itemize}
    \renewcommand{\labelitemi}{\ding{72}}

    \item \textit{Trends}: Integration of HITL within FL remains a novel field of study, lacking a well established methodological framework. Current research focuses primarily on the implementation of HITL mechanisms at the client level, enabling human influence on local models. This approach has demonstrated success in producing more robust and less biased local models~\cite{hirai2023federated, holzinger2023human, abdulrahman2020survey}. However, the black box nature of the FL server poses considerable challenges for implementing HITL mechanisms at this central level, often necessitating additional assumptions. For example, \cite{hausleitner2024collaborative} proposes the generation of synthetic data for subsequent human supervision and labeling, which can then be utilized to fine-tune the global model. A common thread across these studies is the emphasis on developing intuitive user interfaces (UIs) to facilitate effective human interaction with the system, making the HITL application feasible.

\ph{In agentic FL systems, human oversight must extend beyond approving model outputs to supervising autonomous control-plane decisions, such as client participation, objective updates, evaluation gating, or retraining triggers, since these decisions may be taken proactively by agents and can have system-wide implications for trustworthiness~\cite{SAPKOTA2026103599,raza2025trism}.}

    \item \textit{To do}: While client side HITL implementations show promise in aligning models with human preferences, the pervasive issues of data and system heterogeneity in FL present significant obstacles to fully realizing desired outcomes~\cite{hirai2023federated}. Future research should address these challenges, particularly given their prevalence in practical FL system deployments. This may involve exploring novel HITL mechanisms on the server side or using more traditional methods to mitigate client drift.
\end{itemize}

\begin{minorenv}
\begin{enumerate}[label=\textbf{RQ1.1:}, leftmargin=*]
  \item How can HITL in FL be made scalable and resilient to heterogeneity and adversarial manipulation, while keeping oversight decisions auditable and reproducible?
\end{enumerate} 
\end{minorenv}

\paragraph{\textbf{Challenge 1.2: Audience's understanding}} Audience's understanding is a cornerstone for developing \paco{TAI}. Beyond its inherent importance, it partially underpins the crucial requirement for Human Agency and Oversight. To facilitate informed human decision-making, users must be able to effectively query and rectify system behaviors, thus understanding the mechanisms behind model operations and decisions. Although a broader discussion of understanding is provided in Section~\ref{sec:transparency}, this section focuses on aspects of audience's understanding that specifically support human oversight in FL.

\begin{itemize}
    \renewcommand{\labelitemi}{\ding{72}}
    \item \textit{Done:} The interplay between FL and audience's understanding has been extensively investigated since FL's inception. Early research predominantly focused on two key aspects for supporting human oversight: providing explanations to users to enable them to better question and adapt to model outputs and offering mechanisms to trace model behavior back to specific clients and updates. Examples of the former include employing inherently interpretable models~\cite{barcena2022fed}, generating general data prototypes~\cite{pedrycz2021design}, or utilizing established explainability techniques~\cite{chaddad2022explainable}. Meanwhile, approaches for the latter generally involve using distributed ledger technologies, such as Blockchain, to track and verify model updates~\cite{preuveneers2018chained}.

    \item \textit{Trends:} Current research trends in audience's understanding and FL for human oversight show a significant surge, primarily focusing on implementing interpretable models by design, such as trees~\cite{argente2025interpretable, heiyan2024decision, corcuera2025increasing} and rules~\cite{haffar2024glor} based models. Currently, research on Blockchain integration remains active~\cite{bandara2022bassa, kalapaaking2025auditable}, largely driven by the emergence of new communication technologies such as 5G (and the nascent exploration of 6G~\cite{roy2023joint}), which encourage the development of more network efficient solutions for FL.

    \item \textit{To do:} Despite these advancements, several studies highlight the pressing need to enhance support for deep learning models. Additionally, most existing work, with the exception of Blockchain based approaches, primarily considers the classical client-server FL architecture. This underscores the necessity for further research into other practical FL architectures that may be encountered in real world deployments.
\end{itemize}

\begin{minorenv}
\begin{enumerate}[label=\textbf{RQ1.2:}, leftmargin=*]
  \item How can FL systems provide role-specific, privacy-preserving explanations and reporting that enable meaningful oversight without leaking sensitive client or population information?
\end{enumerate} 
\end{minorenv}

\paragraph{\textbf{Challenge 1.3: Inference in Vertical FL}} In Vertical FL (VFL) systems, clients share data samples, but retain their respective labels and features~\cite{rodriguez2023survey}. A representative scenario involves the training of the collaborative model between different entities, such as an insurance company and a bank. A unique challenge arises during the inference phase, where model predictions are also generated in a federated manner. Each client model independently produces an output, which is then aggregated by a learning coordinator (e.g.: a central server) to form a final output. This federated inference process introduces a novel problem for human oversight, creating a ``double black box" problem: both the individual client outputs and their subsequent aggregation remain opaque. To our knowledge, there is no known research that specifically addresses this scenario or proposes mechanisms to enhance human decision-making in such contexts.

\begin{minorenv}
\begin{enumerate}[label=\textbf{RQ1.3:}, leftmargin=*]
  \item How can we make federated inference in VFL interpretable and contestable end-to-end, addressing the ``double black-box'' of client outputs and aggregation?
\end{enumerate} 
\end{minorenv}

\paragraph{\textbf{\ph{Challenge 1.4: Enforcing Autonomy Boundaries and Oversight Gates}} }
\ph{In agentic FL, the degree of autonomy delegated to agents determines the system’s risk profile and its ability to satisfy oversight requirements. While autonomy can improve scalability and responsiveness, uncontrolled autonomy can produce irreversible trust failures through premature deployment, inappropriate policy updates, or unsafe adaptation under drift. Section~\ref{sec:Agentic_Dynamic} motivates autonomy levels (A0-A3) and highlights the need for guardrails; the challenge is to make those guardrails enforceable, verifiable, and usable across multi-stakeholder deployments. }

\ph{In particular, autonomy boundaries should be specified and enforced with respect to explicit autonomy levels (A0-A3), ranging from fully manual control (A0) to policy-constrained execution (A2) and end-to-end autonomous adaptation (A3), since the required oversight, escalation, and evidence obligations differ substantially across these levels.}

\begin{minorenv}
\begin{itemize}
    \renewcommand{\labelitemi}{\ding{72}}
    \item \textit{Done: } Surveys and tutorials on agentic AI provide taxonomies of autonomous capabilities and emphasize oversight as a core requirement for safe operation~\cite{acharya2025agentic,jiang2025large}. TRiSM frameworks formalize trust/risk/security management and reinforce the need for policy constraints and lifecycle governance in agentic AI systems~\cite{raza2025trism}.
    \item \textit{Trends: } Emerging agentic FL visions and decentralized collaboration proposals increasingly treat governance and verifiability as design targets, motivating explicit policy layers and escalation mechanisms~\cite{nguyen2025vision,sarker2025advancing,li2025position}. 
    \item \textit{To do: } A key open problem is to define machine-checkable autonomy policies for each autonomy level (A0-A3), including which decision loci may be automated, what constraints apply, and what escalation gates and audit evidence are required for high-impact actions such as release/rollback, objective changes, and privacy budget updates. The Trust Report~2.0 should encode both autonomy policies and observed policy violations, but standardized representations are still missing.
\end{itemize}
\end{minorenv}

\begin{minorenv}
\begin{enumerate}[label=\textbf{RQ1.4:}, leftmargin=*]
  \item What policy specification and enforcement mechanisms can reliably constrain autonomy in agentic FL (across A0-A3) while remaining auditable and interoperable across multi-stakeholder deployments?
\end{enumerate}
\end{minorenv}

\subsection{Requirement 2: Technical robustness and safety}

The second requirement of \paco{TAI}, technical robustness and safety \cite{diaz2023connecting}, refers to the system’s ability to function reliably and securely under both expected and unforeseen conditions. This includes resilience to attacks, reliability, accuracy, fallback procedures, and reproducibility. Robust AI systems must be able to withstand errors, adversarial behavior, and distributional shifts, ensuring safe and dependable operation throughout their lifecycle. This is especially critical in dynamic or high-risk environments, where failures may have significant consequences.

In the context of FL, achieving technical robustness and safety presents a distinct set of challenges due to the decentralized and heterogeneous nature of the system. In the following, we examine the key challenges that must be addressed to align FL systems with this requirement.

\ph{Beyond classical adversarial threats such as poisoning or Byzantine behavior, agentic FL systems introduce additional failure modes related to goal misalignment, coordination breakdowns, and unintended emergent behavior arising from autonomous decision-making, which are not adequately captured by traditional robustness metrics~\cite{SAPKOTA2026103599,tang2026rethinking}.}

\paragraph{\textbf{Challenge 2.1: Poisoning attacks}} Poison attacks in FL \cite{xia2023poisoning} involve the deliberate injection of malicious data or corrupted model updates by adversarial clients to compromise the integrity of the global model. These attacks present substantial threats \cite{rodriguez2023survey}, particularly within decentralized environments where individual clients maintain control over their local datasets and model updates. Such attacks can severely degrade the performance and reliability of machine learning models, especially when the assumption of independent and identically distributed (IID) data is violated \cite{singh2023fair}.

\begin{itemize}
    \renewcommand{\labelitemi}{\ding{72}}
    \item \textit{Done:} Investigating poisoning attacks has been a cornerstone of FL security research. A significant portion of the early work focused on developing defense mechanisms, primarily at the server level, to counteract these attacks~\cite{sikandar2023detailed}. These defenses often involve robust aggregation techniques that minimize the influence of malicious updates and outlier detection methods applied to model updates~\cite{yazdinejad2024robust, pillutla2022robust, zhang2024anomaly, cristiano2024novel}. In addition, the proposal of novel attack strategies has been extensively explored~\cite{bhagoji2019analyzing}. 
    \item \textit{Trends:} The continuous emergence of novel attack and defense mechanisms has fostered an ongoing "`cat-and-mouse" dynamic within FL security. It is common to observe the proposal of attacks specifically designed to circumvent existing defenses \cite{baruch2019little}, as well as defenses tailored to protect against recently developed "state-of-the-art" attacks \cite{xu2022signguard}, even if their performance against more conventional or simpler mechanisms may vary.

    \item \textit{To do:} Despite substantial advances, considerable challenges remain in the development of defense mechanisms that are both efficient and computationally efficient. Ongoing research continues to prioritize the enhancement of these defense mechanisms to fortify FL systems against the increasing sophistication of poisoning attacks \cite{sharma2024review, almutairi2023federated}, keeping with the trend of the ``cat-and-mouse" game.
\end{itemize}

\begin{minorenv}
\begin{enumerate}[label=\textbf{RQ2.1:}, leftmargin=*]
  \item What defenses can provide lightweight, composition-aware robustness against adaptive poisoning under non-IID data and privacy constraints, with measurable guarantees?
\end{enumerate} 
\end{minorenv}

\paragraph{\textbf{Challenge 2.2: Free riders}}
Free rider attacks in FL \cite{chen2024free} occur when clients participate in the collaborative training process without contributing their local data, with the goal of benefiting from the global model without incurring the associated costs. These attacks can degrade the model's performance and compromise the fairness of the FL system. Defending against free rider attacks is essential to maintain the integrity and effectiveness of the collaborative model training process \cite{lewis2023attacks}. Such defenses aim to ensure that all participating clients contribute meaningfully, thereby preserving the quality and reliability of the global model.
\begin{itemize}
    \renewcommand{\labelitemi}{\ding{72}}
    \item \textit{Done:} The investigation into free rider attacks in FL commenced shortly after FL's introduction. Initial research primarily concentrated on detecting free rider attacks during the initialization phase and the initial communication rounds. This was achieved through various methods, including mutual evaluation mechanisms between clients using ledger technologies such as blockchain \cite{lyu2020towards}, server side anomaly detection \cite{lin2019free}, or reputation mechanisms \cite{xu2020reputation}. \ph{Recent work has begun to formalize trustworthiness under explicitly untrusted participants, emphasizing that robust coordination must be compatible with privacy constraints and realistic client behavior models rather than relying on idealized trusted-client assumptions~\cite{allouah2025untrusted}.}  It is also notable that much of this foundational work in the field often addressed robust aggregation (e.g., to mitigate poisoning attacks) concurrently with the free rider attack problem, leading to more generalized approaches.

    \item \textit{Trends:} With the emergence of more sophisticated free rider attacks \cite{chen2024free}, the literature has increasingly focused on this specific problem, resulting in a divergence between robust aggregation and free rider attack mitigation. Although some previously mentioned methods, such as reputation mechanisms, continue to be explored \cite{nuannimoi2023hyperfed}, recent works exhibit a clear preference for alternative solutions. These include incentive mechanisms \cite{chen2024toward, nguyen2024stake}, which encourage client participation over malicious behavior, and even inference attacks \cite{recasens2024frida} aimed at detecting anomalous and potentially malicious client data distributions, which can indicate a free rider attack.
    
    \item \textit{To do:} Despite the notable evolution of this research area, several challenges persist. These include effectively addressing heterogeneous data and system distributions across clients, which can lead to less accurate predictions of free rider attacks. Additionally, privacy remains a concern, as approaches like those in \cite{recasens2024frida} might compromise the overarching privacy goal of FL. Finally, scalability continues to be a limitation for these methods, suggesting a future research focus on efficiency.
\end{itemize}

\begin{minorenv}
\begin{enumerate}[label=\textbf{RQ2.2:}, leftmargin=*]
  \item How can FL detect and filter out free riders reliably without compromising privacy, incentives, or fairness under partial participation and heterogeneous clients?
\end{enumerate} 
\end{minorenv}

\paragraph{\textbf{Challenge 2.3: Out Of Distribution Detection}} Out-of-Distribution (OOD) detection has emerged as a critical area of research in machine learning, primarily due to the inherent difficulty of models in generalizing effectively to data significantly different from their training distribution. Such divergent data can lead to erroneous yet confident predictions, potentially resulting in unreliable and dangerous outcomes. Consequently, identifying when the input data deviates from the training distribution is crucial to ensure the safety and reliability of the model~\cite{yang2024generalized}. The complexities of OOD detection are further aggravated within FL environments. The decentralized nature of FL, characterized by Non-IID data between participating clients and restricted access to individual training samples due to privacy constraints, significantly challenges the applicability of many effective OOD detection techniques, rendering this field particularly challenging~\cite{guo2023out}.

\begin{itemize}
    \renewcommand{\labelitemi}{\ding{72}}
    \item \textit{Done:} OOD detection is a widely researched problem within FL. Initial efforts mainly focused on adapting OOD detection mechanisms from centralized learning to federated environments. This often involved training anomaly detection models such as LSTMs or GRUs \cite{sater2021federated, mothukuri2022federated}, implementing data augmentation at the client level \cite{weinger2020enhancing}, or training a model on a centralized known dataset before federating it \cite{raza2022using}. Although these techniques have shown some results, their interplay with FL has not been proven to be perfect.

    \item \textit{Trends:} More recent approaches distinguish themselves from earlier work by introducing novel techniques not previously employed in centralized learning, usually by exploiting properties inherent to federated settings, resulting in more innovative and efficient solutions. Examples include FOODG \cite{liao2024foogd}, a framework that integrates OOD detection with OOD generalization. This method trains a federated score matching model and utilizes regularization in the local loss function to better align the models for improved detection. Another example is Fin-Fed-OD \cite{herurkar2024fin}, which leverages data distribution shifts between clients by comparing latent representations derived from client-owned autoencoders. \major{Nevertheless, none of this approaches provides an optimal solution. The need for privacy preserving OOD detection methods has been lacking in literature~\cite{jeong2025out}. Furthermore, other works also missed the tools for a validating OOD detection in real world settings where the nature of this problem can widely vary~\cite{herurkar2024fin}.}
    
    \item \textit{To do:} Although FL was developed with privacy in mind, current research often overlooks whether the OOD detection techniques used might inadvertently leak sensitive information \cite{liao2024foogd}. This requires a comprehensive evaluation from an adversarial perspective.
\end{itemize}

\begin{minorenv}
\begin{enumerate}[label=\textbf{RQ2.3:}, leftmargin=*]
  \item How can FL perform reliable OOD detection and safe adaptation using only decentralized signals, while avoiding new attack surfaces and privacy leakage?
\end{enumerate} 
\end{minorenv}

\paragraph{\textbf{\ph{Challenge 2.4: Evaluation Integrity under Agentic AI Coordination}}} 
\ph{Evaluation integrity refers to ensuring that model assessment in federated systems remains reliable, representative, and resistant to manipulation across time. Agentic FL introduces new risks: autonomous components can select metrics, curate test suites, gate evaluations, or suppress regressions as part of control-plane optimization. In LLM-enabled workflows, evaluation may also depend on semantic judgments, tool invocation, or dynamic benchmark selection, increasing the attack surface. These risks align with the agency-specific threats identified in Section~\ref{sec:Agentic_Dynamic}, particularly evaluation manipulation and metric gaming, and they directly impact robustness, fairness, and safety. }

\ph{It is important to point out that this challenge is not about improving evaluation metrics, but about preserving the integrity and governance of evaluation under autonomous control.}

\begin{minorenv}
\begin{itemize}
    \renewcommand{\labelitemi}{\ding{72}}
    \item \textit{Done: } The agentic AI and LLM trustworthiness literature explicitly characterizes weak-to-strong trust issues and the risk of optimizing proxy objectives, reinforcing the need for robust evaluation protocols~\cite{sun2025generalizing}. Secure semantic communication perspectives further highlight how generative/agentic systems reshape threat models, including integrity risks at the semantic layer~\cite{tang2026rethinking}.
    \item \textit{Trends: } TRiSM-oriented approaches emphasize lifecycle controls and risk governance for LLM-based multi-agent systems, including evaluation, monitoring, and guardrails~\cite{raza2025trism,zhang2026security}. In federated reasoning and distributed LLM surveys, evaluation challenges are increasingly discussed as system-level concerns in heterogeneous and distributed settings~\cite{wei2025federated,amini2025distributed}.
    \item \textit{To do: } Key open problems include defining evaluation governance primitives (who selects metrics, who approves changes, what triggers a rollback), hardening evaluation pipelines against adaptive manipulation, and designing drift-aware evaluation suites that remain valid under temporal non-stationarity. Standardized “evaluation provenance” fields suitable for Trust Report~2.0 remain underdeveloped.
\end{itemize}
\end{minorenv}

\begin{minorenv}
\begin{enumerate}[label=\textbf{RQ2.4:}, leftmargin=*]
  \item How can agentic FL enforce evaluation integrity (metric stability, test provenance, drift-aware validation) when control-plane components can adapt objectives and gate or modify evaluations over time?
\end{enumerate}
\end{minorenv}

\subsection{Requirement 3: Privacy and data governance}

The third requirement of \paco{TAI}, privacy and data governance, emphasizes responsible handling of personal and sensitive data throughout the lifecycle of the AI system. This involves ensuring data protection, enabling secure data processing, and providing individuals with meaningful control over their information. FL directly supports this objective by enabling model training without the need to centralize raw data, thus mitigating the risk of data exposure. Among the \paco{TAI} requirements, this area is often considered less challenging, as FL was explicitly developed with the preservation of privacy in mind. Consequently, less foundational adaptation is required compared to other domains. However, FL does not eliminate all privacy risks. Model updates can still leak sensitive information, and the implementation of secure aggregation, effective data governance, and regulatory compliance remains a significant challenge in decentralized settings. The following section elaborates on these specific challenges.

\paragraph{\textbf{Challenge 3.1: Inference attacks}} Inference attacks in FL \cite{rao2024privacy} pose a significant threat to privacy by allowing the adversarial clients to extract sensitive information about individual clients' data from shared model updates. These attacks exploit the fact that model updates, even when aggregated, may still contain patterns that can be reverse-engineered to infer private data. A notable example is the potential to infer characteristics about a client's data through model weights, gradients, or output predictions shared during the federated training process \cite{yang2023practical}.

\begin{itemize}
    \renewcommand{\labelitemi}{\ding{72}}
    \item \textit{Done:} Although FL was designed with user privacy in mind, research soon demonstrated the necessity of additional defensive measures. Early work proved that client data could be reconstructed through shared gradients or model updates~\cite{zhu2019deep, geiping2020inverting}. Additionally, membership inference attacks became prevalent, enabling adversaries to deduce whether a specific client participated in the training of a given model \cite{zhao2021user}. Differential Privacy (DP) emerged as the most common defense, involving the addition of noise to model updates before they are transmitted to the server \cite{geyer2017differentially}. Alternative approaches to DP such as Secure Multi-Party Computation (SMPC) \cite{liu2024survey} and Homomorphic Encryption (HE) \cite{xie2024efficiency} were also broadly explored.

    \item \textit{Trends:} Current works focus on improving scalability of already existing solutions in order to make their application to the real world settings feasible~\cite{xie2024efficiency}. Furthermore, studies are becoming more narrow in terms of their applicability, introducing works which focus on specific FL scenarios such as cross-silo FL or cross-device FL~\cite{sen2025privacy}. Finally, recent proposals combine multiple technologies and suggest multi-layered approaches for addressing the inference attacks problem, such as studying the interplay of DP and HE, its advantages and limitations~\cite{sen2025privacy}. \major{Nevertheless, these efforts are not without shortcomings. Current solutions have an inherent trade-off among computational efficiency, model performance, and security against state-of-the-art attacks rendering them not yet ideal for TAI systems.}

    \item \textit{To do:} The application of DP offers enhanced privacy at the cost of a significant reduction in performance. This represents the most critical open challenge in this field. Furthermore, with the recent adoption of techniques like SMPC which requires more resources, computational efficiency has also become an important challenge in the current literature \cite{liu2024survey}. Finally, the intersection between robust and privacy-aware training is a promising research area, seeking for a method that is able to protect from both poisoning and inference attacks~\cite{xu2024dual}.
\end{itemize}

\begin{minorenv}
\begin{enumerate}[label=\textbf{RQ3.1:}, leftmargin=*]
  \item How can we characterize and mitigate multi-channel inference risk in FL beyond DP parameters, under realistic auxiliary information and evolving threat models?
\end{enumerate} 
\end{minorenv}

\paragraph{\textbf{Challenge 3.2: Model Unlearning}} Model unlearning~\cite{bourtoule2021machine} involves the removal of the influence of specific data points or concepts from a trained machine learning model, ideally without compromising the model's overall performance. This field has gathered significant attention, largely driven by regulatory frameworks such as the GDPR~\cite{gdpr2} and the ``right to be forgotten", which grant users the ability to request the removal of their personal data. Should such data have been utilized in model training, model unlearning becomes a necessary procedure. However, this challenge is even more difficult in federated environments. Here, the objective is to eliminate the influence of a particular client across multiple training rounds, a task made considerably more complicated by the absence of direct access to any data that could serve as a reference for the unlearning process.
\begin{itemize}
    \renewcommand{\labelitemi}{\ding{72}}
    \item \textit{Done:} Early approaches in the literature concerning model unlearning in FL often necessitated maintaining a historical record of parameters and model updates~\cite{liu2020federated, yan2023federated}. These significant requirements, which extended even after model deployment, requiring alternative research directions. Some works imposed specific restrictions on models, such as those employing Bayesian Variational Inference~\cite{gong2021bayesian}, to facilitate easier unlearning. While these methods proved efficient, they often introduced restrictive mechanisms.

    \item \textit{Trends:} With a strong emphasis on efficiency, current research in this domain explores diverse approaches, frequently drawing inspiration from other fields. For instance, the use of adapters has recently gained traction, influenced by the considerable interest in model merging techniques~\cite{zhong2025unlearning}. Similarly, disentangling client contributions, a concept rooted in representation learning, is being investigated~\cite{wang2024learning}. Furthermore, parameter selection via explanations, reflecting the growing field of explainability, is also showing promise~\cite{xu2025update}. \major{However, current approaches show significant drawbacks, such as including a significant computational and communication overhead~\cite{xu2025update}, degrading model performance~\cite{varshney2025unlearning} or the lack of theoretical results on the field~\cite{huynh2024fast}, failing to provide an optimal solution to this challenge.}
    
    \item \textit{To do:} The field is expected to continue its pursuit of more efficient unlearning mechanisms, focusing on both computational resources and model performance. Observing current trends, there is no single methodology that has emerged as a clear standard, suggesting that future work will likely involve a variety of creative and innovative approaches \major{which may include the development of theoretical results}.
\end{itemize}

\begin{enumerate}[label=\textbf{RQ3.2:}, leftmargin=*]
  \item What does verifiable and effective unlearning mean in FL across rounds and clients, and which guarantees are achievable without prohibitive retraining and communication cost?
\end{enumerate} 

\paragraph{\textbf{\ph{Challenge 3.3: Semantic-Layer Leakage and Orchestration Risk in LLM-Enabled FL}}}
\ph{LLM-enabled federated systems often introduce a semantic layer in which agents exchange structured natural language, high-level summaries, or tool-mediated signals to coordinate decisions. While such semantic communications can improve efficiency and flexibility, they introduce new privacy and security risks that differ from classical gradient leakage, including unintended disclosure through semantic summaries, prompt-driven leakage, or toolchain-mediated exfiltration. These risks become particularly acute in edge or multi-modal deployments, where constraints and heterogeneous modalities further complicate assurance.}

\ph{This challenge does not aim to comprehensively address LLM security, but focuses specifically on semantic exchanges used for federated coordination.}

\begin{minorenv}
\begin{itemize}
    \renewcommand{\labelitemi}{\ding{72}}
    \item \textit{Done: } Work on secure semantic communications emphasizes new threats and opportunities introduced by generative and agentic systems, motivating security models that extend beyond parameter-level protection~\cite{tang2026rethinking}. Surveys on distributed LLMs and edge deployment highlight system constraints and attack surfaces that emerge in practical deployments~\cite{amini2025distributed,kristiani2026deploying}.
    \item \textit{Trends: } Federated reasoning LLMs and multi-agent communication fabrics indicate increased use of higher-level coordination abstractions, including semantics-aware communication, which increases the importance of semantic privacy and integrity controls~\cite{wei2025federated,giusti2025federation}.
    \item \textit{To do: } Key open problems include defining privacy notions and defenses for semantic exchanges (beyond gradient-level leakage), developing auditing interfaces for LLM-mediated orchestration, and integrating semantic-layer risk signals into Trust Report~2.0. Establishing threat models that capture prompt/tool injection and semantic exfiltration without assuming centralized inspection remains a major challenge.
\end{itemize}
\end{minorenv}

\begin{minorenv}
\begin{enumerate}[label=\textbf{RQ3.3:}, leftmargin=*]
  \item How can federated systems protect and audit semantic-layer communications and LLM-mediated orchestration, ensuring privacy and integrity without sacrificing decentralization or operational feasibility?
\end{enumerate}
\end{minorenv}

\subsection{Requirement 4: Transparency}\label{sec:transparency}

The fourth requirement of \paco{TAI}, transparency, refers to the need for AI systems to be understandable, traceable, and communicable to all relevant stakeholders. This involves clearly documenting system capabilities and limitations, ensuring the traceability of decisions, and enabling meaningful explanations, especially in contexts where outputs have significant consequences.

In FL, achieving transparency is particularly challenging due to the decentralized nature of the system, the lack of visibility into client-side data and processes, and the complexity of coordinating updates across a distributed network. These factors make it more difficult to trace model behavior, communicate rationale, and assess accountability, although doing so is essential to build user Trust and ensure responsible deployment.

\paragraph{\textbf{Challenge 4.1: eXplainable Artificial Intelligence}} A key challenge for transparency in FL lies in the limited explainability of models trained in decentralized and heterogeneous environments. This issue becomes more critical when dealing with complex architectures, such as deep neural networks, whose decision-making processes are inherently opaque \cite{herrera2025reflections}. In FL, the lack of access to raw client-data and the variability of local contexts further hinder efforts to generate consistent and interpretable explanations across clients.

\begin{itemize}
    \renewcommand{\labelitemi}{\ding{72}}
    \item \textit{Done:} Early researches on explainability in FL often adapt existing proposals to FL employing post-hoc explanations for already trained models \cite{ben2022trust}. While this was a dominant trend, some works began to address aspects inherent to the federated schema, such as developing interfaces for coordinating the training process and presenting behavior in an interpretable manner for the end-user \cite{wei2019multi}.

    \item \textit{Trends:} Current research leverages explainability methods and the federated nature of FL to enhance the final model, which represents a paradigm shift from previous approaches. A representative example of this shift of paradigm is using feature relevance to appropriately weight clients during aggregation \cite{ali2025explainable}. Another notable example involves quantifying uncertainty at the client level to achieve a more accurate global estimate and improved predictions \cite{zhang2025uncertainty}. Nonetheless, some ongoing work still focuses on generating post-hoc explanations in FL \cite{kalakoti2024explainable}.
    
    \item \textit{To do:} Moving forward, future directions include developing 
    XAI frameworks specifically tailored for heterogeneous client configurations. Additionally, there's a need for protocols for federated auditing and explanation alignment. Bridging the gap between technical explanations and human interpretability, particularly in low-resource or non-expert environments, will be crucial for maintaining transparency at scale in FL deployments.
\end{itemize}

\begin{minorenv}
\begin{enumerate}[label=\textbf{RQ4.1:}, leftmargin=*]
  \item How can we deliver explanations that are faithful, stable under non-IID and drift, and consistent across clients, without introducing new privacy or security vulnerabilities?
\end{enumerate} 
\end{minorenv}

\paragraph{\textbf{Challenge 4.2: Causal Artificial Intelligence}} Causal Artificial Intelligence (CAI) is gaining increasing attention in AI \cite{rawal2025causality} as a powerful tool for uncovering the underlying mechanisms of data generation, allowing robust generalization beyond correlations. Unlike traditional statistical models, causal models aim to capture invariant relationships that remain stable across interventions and domain shifts. Incorporating causal reasoning into FL \cite{kayaalp2024causal} holds the promise of more reliable, interpretable, and robust decentralized models. In particular, causal AI can help FL systems learn stable features across heterogeneous clients \cite{xiong2023federated}, improve out-of-distribution performance, and support counterfactual reasoning \cite{hasan2024counterfactual} for downstream tasks such as personalized treatment recommendation, fairness analysis, or domain adaptation.

However, the integration of causal inference methods into FL poses unique and underexplored challenges \cite{kayaalp2024causal}. First, causal discovery and estimation typically require access to rich, interventional, or diverse observational data, something that is difficult       to guaranty in distributed, privacy sensitive clients \cite{vo2022bayesian}. Clients may have partial, biased, or structurally different data distributions, which complicates the identification of shared causal structures. Moreover, coordinating causal assumptions or graphical models between clients without data centralization raises questions about the consistency, identifiability, and validity of the model \cite{ye2024federated}. The non-IID nature of federated data further aggravates the risk of learning spurious or unstable causal relationships when pooling gradients or model updates.

\begin{itemize}
    \renewcommand{\labelitemi}{\ding{72}}
    \item \textit{Trends:} The study of the interplay between CAI and FL is relatively new, leading to the concurrent exploration of several research directions \cite{wang2023towards}. One promising approach is learning causal representations in FL \cite{tang2023learning}, where shared representations aim to encode causal factors while filtering out spurious correlations; this area is often also referred to as OOD generalization \cite{liao2024foogd}. Furthermore, some studies propose federated variants of causal discovery algorithms using techniques such as decentralized constraint optimization, Blockchain, or SMPC to maintain privacy \cite{mu2024explainable}. Finally, integration of domain knowledge or structural priors at the client level shows promise in guiding the FL training process toward more causally sound inferences \cite{yang2024federated}.

    \item \textit{To do:} The intersection of causality and FL represents a nascent research field, and consequently a significant volume of work is expected in the coming years. The primary challenge that requires attention is the heterogeneity of the data at the client level, which often hinders proper causal structural learning in numerous scenarios. In addition, scalability and communication efficiency emerge as relevant challenges within this domain~\cite{mu2024explainable}.
\end{itemize}

\begin{minorenv}
\begin{enumerate}[label=\textbf{RQ4.2:}, leftmargin=*]
  \item How can causal claims in FL be estimated and communicated with explicit assumptions and uncertainty when data and interventions are distributed and partially unobservable?
\end{enumerate} 
\end{minorenv}

\paragraph{\textbf{Challenge 4.3: Models Interpretable by Design}} The adoption of intrinsically more interpretable models has gained significant traction, particularly within high-risk domains such as finance and healthcare~\cite{ducange2024federated}. In contrast to opaque black-box models, inherently interpretable models, including decision trees, linear models, and rule-based systems, provide intuitive insights into the relationship between inputs and outputs, empowering users to understand, validate, and challenge predictions. However, within FL, the inherent non-IID nature of the data between clients presents a considerable challenge to the generalizability of these simpler models~\cite{argente2025interpretable}. Furthermore, certain models, such as decision trees, may require the transmission of symbolic information rather than conventional gradient or model updates~\cite{daole2025Trustworthy}, thus requiring specific adaptations to effectively integrate them into the FL framework.

\begin{itemize}
    \renewcommand{\labelitemi}{\ding{72}}
    
    \item \textit{Done:} Earlier research mainly focused on methods for federated training of non gradient descent models, such as random forests or decision rules, with the goal of preserving privacy \cite{chen2021fed, qiao2021learning}. This scientific exploration was somewhat limited, largely driven by the non-immediate applicability of these methods in practical FL deployments.

    \item \textit{Trends:} Current work is beginning to view these types of model as a source of explainability and a way to satisfy the transparency requirements. New research directions address practical challenges encountered in real-world FL deployments, including data heterogeneity \cite{daole2025Trustworthy} and system-level considerations \cite{lin2020ensemble}. Furthermore, the application of this research area to various scenarios, such as healthcare care and forensics, is gaining traction \cite{de2022interpretable, chaddad2022explainable}. However, despite this new research direction, previous ones, such as the adaptation of specific models to the FL paradigm, remain an active area of study \cite{haffar2024glor}.
    
    \item \textit{To do:} Future work must prioritize the implementation of more advanced privacy mechanisms in training these systems, especially given the proliferation of new privacy attacks in federated settings \cite{de2022interpretable}. Furthermore, while heterogeneity has begun to be explored, it remains one of the most significant challenges in this field \cite{daole2025Trustworthy} and requires substantial further investigation due to the significant performance loss under these scenarios.
\end{itemize}

\begin{minorenv}
\begin{enumerate}[label=\textbf{RQ4.3:}, leftmargin=*]
  \item Which inherently interpretable model classes and training protocols remain competitive under FL heterogeneity while respecting privacy constraints and deployment realities?
\end{enumerate} 
\end{minorenv}

\paragraph{\textbf{Challenge 4.4: Data Provenance}} Data provenance, documenting the origin and lifecycle of data used to train a model, including the collection context and transformations~\cite{souza2019provenance}, has become indispensable for transparency because it enables users, auditors and stakeholders to assess data quality, bias sources, and error chains. In FL, however, maintaining a unified provenance trail is difficult: privacy-by-design prevents centralizing raw records; client-side logs may be heterogeneous; and dynamic participation (clients joining/leaving) breaks continuity. \major{Rather than a single central ledger, provenance in FL should be privacy-thrifty and verifiable: clients keep tamper-evident, hash-chained local logs; the server stores only pseudonymous, signed update receipts and policy-compliance proofs (e.g., DP budget ledgers, acceptance criteria satisfied); and any cross-node reporting is limited to aggregated or DP-protected summaries. A complementary system view is to implement provenance with a database-centric layer for lineage capture, standardized schemas, and auditable queries, while enforcing privacy constraints at query time \citep{gu2024enhancing}. This shifts transparency from exposing data to exposing guarantees (who did what, under which policies, with what privacy budget), preserving traceability for audits while avoiding re-identification and metadata leakage.}

\begin{itemize}
    \item \textit{Trends:} Interest in the provenance of data within FL is a relatively recent development. Some research explores the use of distributed ledger technologies, such as blockchain, to track model updates down to the client level \cite{preuveneers2018chained, bandara2022bassa, nourmohammadi2023privacy}. However, these ledger technologies can add architectural complexity, leading to the consideration of alternative approaches. For example, TraceFL \cite{gill2023tracefl} introduces a novel mechanism to track contributions at the parameter level of the model, enabling fine-grained control. Another prominent approach involves the use of watermarking methods for deep neural networks to efficiently track contributions \cite{lansari2023federated}. Furthermore, integration of zero-knowledge proofs (ZKPs), a cryptographic method that allows one party to prove to another that a statement is true without revealing any information beyond the validity of the statement itself, is being studied, showing promising results \cite{li2025integrating}.

    \item \textit{To do:} Future work in this area will be largely driven by advances in related research fields. For example, progress in neural network watermarking~\cite{lansari2023federated}, itself a nascent field, could significantly improve the provenance capabilities of the data. Similarly, the integration of ZKPs with machine learning is an active area of research that has considerable potential. Furthermore, investigating existing methods from an adversarial perspective could yield crucial insights into their viability and whether they expose sensitive information about the federated training process. Future work should quantify provenance completeness, proof latency/size (for ZKPs), and watermark robustness (under pruning/quantization), alongside explicit privacy-leakage bounds for provenance artifacts.
    
\end{itemize}

\begin{minorenv}
\begin{enumerate}[label=\textbf{RQ4.4:}, leftmargin=*]
  \item How can FL provide verifiable, privacy-thrifty provenance with \- bounded leakage and acceptable overhead at scale?
\end{enumerate} 
\end{minorenv}

\subsection{Requirement 5: Diversity, non-discrimination \& fairness}

The fifth requirement of \paco{TAI}, diversity, nondiscrimination, and fairness, emphasizes the need for AI systems to treat all individuals and groups equitably while considering the social and cultural contexts in which they operate. This involves avoiding unfair bias, ensuring equal access and representation, and promoting inclusive design practices throughout the entire AI lifecycle. In the context of FL, this requirement poses particular challenges due to the inherent heterogeneity of the data between clients. Variations in demographic representation, data quality, or device capabilities can lead to models that perform poorly for minority or underrepresented groups, reinforcing systemic inequalities. Addressing these issues in FL requires methods that promote fairness between decentralized data sources, while respecting local privacy constraints and maintaining performance parity.

\ph{In agentic and dynamic FL settings, fairness risks may also emerge over time as autonomous adaptation or client-selection policies disproportionately affect certain subpopulations, making temporal monitoring of disparity an essential complement to static fairness assessments.}

\paragraph{\textbf{Challenge 5.1: Data heterogeneity}} Data heterogeneity refers to the phenomenon in which client data distributions diverge significantly from one another within an FL ecosystem. In practical FL deployments, key manifestations of data heterogeneity include: \begin{enumerate*}[label=(\arabic*)]
    \item \textit{Non-IID Data Distributions} \cite{ma2022state}, where client datasets frequently violate the IID assumption. This deviation introduces biases that can detrimentally impact the performance and generalization capabilities of the global model;
    \item \textit{Concept and Covariate Shifts} \cite{huang2023rethinking}, where variations in the underlying feature-label relationships (concept shift) and discrepancies in feature distributions (covariate shift) across clients pose substantial challenges to the model's capacity for effective generalization across the heterogeneous data landscape; and
    \item \textit{Data Quantity Imbalances} \cite{fang2022robust}, where disparities in the volume of data contributed by individual clients can lead to a quantity skew. This imbalance may result in clients possessing larger datasets that disproportionately influence the global model, increasing the risk of overfitting to their specific data characteristics.
\end{enumerate*}

\begin{itemize}
    \renewcommand{\labelitemi}{\ding{72}}
    \item \textit{Done:} Data heterogeneity has consistently presented a significant challenge in FL~\cite{li2020challenges}. Consequently, numerous research efforts emerged soon after the inception of FL to address this issue. Among pioneering works are FedProx \cite{li2020federated}, SCAFFOLD \cite{karimireddy2020scaffold}, and FedNova \cite{wang2020tackling}. These frameworks continue to serve as fundamental baselines for subsequent methodologies in the field. These early works rigorously demonstrate the inefficiencies of FedAvg in optimizing functions under data heterogeneity. They show that FedAvg, in such scenarios, ultimately optimizes a surrogate function rather than the intended objective function. For instance, FedNova illustrates this outcome specifically in the context of data quantity imbalances, while FedProx provides similar proofs for non-IID data distributions and concept shift scenarios.
    \item \textit{Trends:} Among recent work, the use of personalized FL methods is gaining considerable traction \cite{ranaweera2025enhancing}. These approaches aim to produce a global model that can subsequently be adapted to each client's specific data distribution. Currently, client clustering, an approach that groups clients with similar data distributions, is actively being explored \cite{liu2025survey}. This strategy effectively reduces heterogeneity within a given cluster, leading to the generation of distinct models tailored for each cluster. Furthermore, some research efforts are dedicated to developing robust aggregation techniques \cite{pei2024review}, which are capable of mitigating the impact of outlier data, thus enhancing the overall resilience of FL systems.
    \item \textit{To do:} While current research trends in addressing data heterogeneity yield significant results, it is imperative that future work addresses their inherent limitations. For example, personalized FL approaches require that each client have a sufficient volume of local data to effectively adapt the global model to their specific distribution. Although client grouping could potentially mitigate this issue, grouping clients based on data distribution could inadvertently lead to the leakage of sensitive information in certain scenarios. Finally, while robust aggregation techniques are effective in reducing or even ignoring the impact of outlier data, this process can unfortunately result in the loss of useful information that may be crucial for the overall performance of the model in specific applications~\cite{sahoo2024feddual}. Finally, it has been observed that data heterogeneity can lead to poor fairness in the resulting model~\cite{chen2025advances} while fairness optimization leads to a good model generalization~\cite{krueger2021out}, showing that fairness is a prominent tool to address data heterogeneity.
\end{itemize}

\begin{minorenv}
\begin{enumerate}[label=\textbf{RQ5.1:}, leftmargin=*]
  \item How can FL achieve fair and robust performance under non-IID distributions, shifts, and quantity imbalance, especially for small or minority subpopulations, without requiring sensitive attributes centrally?
\end{enumerate} 
\end{minorenv}

\paragraph{\textbf{Challenge 5.2: System heterogeneity}} System heterogeneity in FL encompasses differences in client devices' hardware and network capabilities, introducing several challenges: \begin{enumerate*}[label=(\arabic*)]
    \item \textit{Device Resource Constraints} \cite{imteaj2022federated}, where variations in computational power, memory, and energy availability among clients; 
    \item \textit{Network Connectivity Variability} \cite{yang2022federated}, where inconsistent and limited network access among clients can lead to communication delays and synchronization issues, affecting the timely aggregation of model updates; and
    \item \textit{Model Architecture Diversity} \cite{wang2023towards}, where differences in local model architectures due to hardware limitations or personalized tasks.
\end{enumerate*} 

Addressing these issues requires the implementation of effective \textit{client management} strategies~\cite{shanmugarasa2023systematic, li2024comprehensive}. Client management is crucial for ensuring system robustness (by accommodating diverse client behaviors within the training protocol), promoting fairness (by ensuring equitable representation across various client groups), and optimizing performance (by leveraging contributions from clients with potentially superior model performance due to their unique data distributions). 

\begin{itemize}
    \renewcommand{\labelitemi}{\ding{72}}
    \item \textit{Done:}  Investigations into mitigating the challenge of system heterogeneity emerged shortly after the inception of FL, with initial efforts focusing on client selection strategies~\cite{nishio2019client, goetz2019active}. These strategies aimed primarily to enhance the efficiency of the training process by prioritizing clients with superior model performance or greater computational capabilities. This emphasis on efficiency constituted the predominant research objective during the nascent years of FL.  For example, the already introduced FedNova~\cite{wang2020tackling} framework reframes the problem of device resource constraints as one of data quantity imbalance, positing that the available data correspond to the amount a given device can process within a specified timeframe. Other notable approaches include q-Fair FL~\cite{Li2020Fair}, which introduces a minimization objective designed to ensure comparable accuracy between different devices, thus preventing certain devices from gaining an undue advantage. Another prominent example, using a distinct approach, is a level-based FL~\cite{chai2020tifl}, which classifies clients into multiple levels based on their training performance and selects clients from only a designated tier in each training round.

    \item \textit{Trends:} Recent research extends beyond device resource constraints to explore other critical aspects of system heterogeneity. For example, FedPartial~\cite{jian2025fedpartial} addresses network connectivity variability by enabling model aggregation with only partial client updates. Currently, client selection is under active investigation~\cite{ami2025client}, in order to identify optimal client subsets that reduce training latency while preserving generalization capabilities. Furthermore, sparsity is being explored to allow for variations in model size, thereby facilitating the deployment of more efficient models tailored to the specific computational capabilities of individual clients~\cite{chen2023efficient, liu2024sparse}. Recent academic efforts have also expanded the scope of client management in FL beyond simple efficiency considerations to encompass a broader range of objectives, including fairness~\cite{huang2022stochastic} and mitigation of client dropouts~\cite{jiang2024dordis, wang2022combating}. Currently, research on incentive mechanisms remains an active area of investigation, with distributed ledger technologies, particularly Blockchain, demonstrating considerable promise as tools for client motivation~\cite{behera2021federated, qu2021proof}.
    
    \item \textit{To Do:} Future research directions include new approaches such as an incentive mechanism to keep clients engaged and avoid abandonment of the connection~\cite{chen2025advances}. Future efforts within this research domain should increasingly take into account more realistic scenarios, such as dynamic network conditions, a factor that is largely overlooked in the current literature~\cite{gouissem2024comprehensive}. Furthermore, certain client selection strategies can inadvertently facilitate the leakage of sensitive information, thus contradicting the fundamental tenets of FL. Furthermore, prospective research on incentive mechanisms must address the inherent risks these mechanisms pose to system robustness and security, specifically by preventing their exploitation by malicious actors~\cite{ali2023systematic}.
\end{itemize}

\begin{minorenv}
\begin{enumerate}[label=\textbf{RQ5.2:}, leftmargin=*]
  \item How can FL adapt to heterogeneous network constraints without inducing systematic performance disparities or privacy leakage through participation and selection effects?
\end{enumerate} 
\end{minorenv}

\subsection{Requirement 6: Societal and environmental well-being}

The sixth requirement of \paco{TAI}, societal and environmental well-being, underscores the importance of ensuring that AI systems contribute positively to individuals, communities, and the planet. This includes promoting sustainability, fostering social cohesion, and avoiding adverse impacts on collective well-being. In the context of FL, this requirement takes on a dual dimension. On the one hand, FL has the potential to support socially beneficial applications, such as privacy-preserving healthcare or personalized education, by enabling collaborative learning without centralizing sensitive data. However, the distributed nature of FL can lead to increased energy consumption due to repeated local training and communication, particularly in large-scale deployments or when combined with resource-intensive models like LLM. Ensuring that FL systems align with societal goals while minimizing environmental costs is therefore essential for their responsible and sustainable adoption.

\ph{Agentic operation can further amplify sustainability concerns, as autonomous retraining, evaluation, or orchestration decisions may increase communication frequency and computational overhead unless explicitly constrained by governance policies.}

\paragraph{\textbf{Challenge 6.1: High Communication costs}}
 The high communication costs represent a significant challenge in FL \cite{thomas2025beyond}, primarily due to the frequent transmission of model updates between clients and the central server. This process can be particularly demanding for devices with limited network bandwidth, IoT devices, or smartphones, which are commonly used in FL scenarios \cite{zhang2022federated}. Substantial communication overhead not only strains network resources, but also increases latency, potentially hindering the efficiency of the learning process.

\begin{itemize}
    \renewcommand{\labelitemi}{\ding{72}}
    \item \textit{Done:} FL was designed with communication efficiency in mind~\cite{bib:mcmahan16communicationefficient}, claiming a x10-100 reduction in communication rounds over synchronized SGD. However, several works appeared after FL's introduction which aimed to reduce the communication overhead~\cite{konevcny2016federated}, leveraging tools such as model compression and structured updates, which learn updates from a constrained space with fewer variables. The techniques for model compression vary, with model pruning~\cite{sattler2020robust} being the most common approach. Furthermore, Federated Dropout~\cite{caldas2018expanding}, a technique that allows clients to learn submodels, was also introduced.
    \item \textit{Trends:} The current literature can be broadly categorized into three primary approaches~\cite{le2024exploring}: (1) reducing the number of communication rounds, (2) decreasing the number of participants, and (3) employing model compression techniques. This classification underscores the prevalence of model compression, which was a focus even in earlier works, while showing the adoption of new perspectives. Within the first category, methods such as FedProx~\cite{li2020federated} and FedNova~\cite{wang2020tackling} are notable for their ability to accelerate the generalization of the model, thus addressing the interaction between communication efficiency and data heterogeneity. The second category, which aims to minimize the number of participating clients, involves ongoing exploration of various client selection methodologies, again highlighting the interrelation between this challenge and others previously discussed~\cite{fu2023client}. The third and final category, model compression, has yielded a substantial body of novel and specialized research. Common techniques in this area include model pruning, sparsification, and factorization techniques~\cite{woisetschlager2024survey}.
    
    \ph{Edge-facing deployments in consumer IoT further highlight that "trustworthy" operation must remain feasible under tight resource constraints; optimized FL designs for edge decision-making explicitly foreground latency, compute limits, and reliability considerations that can constrain the achievable trust envelope~\cite{Rehman2025}.}

    \item \textit{To Do:} Future research efforts should prioritize investigating novel paradigms and scenarios, including but not limited to Federated Transfer Learning (FTL) and the exploration of ad-hoc privacy-preserving methodologies. Currently, continued advances in established approaches, such as dynamic client allocation and selection, remain crucial. Ultimately, the establishment of a standardized benchmark is imperative to facilitate a rigorous comparison and analysis of the proposed methods.
\end{itemize}

\begin{minorenv}
\begin{enumerate}[label=\textbf{RQ6.1:}, leftmargin=*]
  \item Which objectives and protocols best reduce communication overhead while preserving privacy, robustness, and model quality in resource-constrained deployments? 
\end{enumerate} 
\end{minorenv}

\paragraph{\textbf{Challenge 6.2: Limited Bandwidth}} Limited bandwidth poses a significant challenge in FL \cite{zhang2022federated}, as the frequent exchange of model updates between clients and the central server can be hindered by network constraints. This issue is particularly pronounced in devices with restricted communication capabilities \cite{ji2023joint}, such as IoT devices and smartphones, which are commonly employed in FL scenarios.

\begin{itemize}
    \renewcommand{\labelitemi}{\ding{72}}
    \item \textit{Done: } This particular challenge exhibits a strong correlation with the issue of high communication costs. Consequently, numerous approaches address both concerns simultaneously, sharing a substantial portion of the initial research efforts. However, specialized work has focused on developing targeted strategies, such as Deep Gradient Compression~\cite{lin2018deep}. This particular method achieves an approximate 270-fold compression ratio without compromising performance, thus establishing itself as a prominent approach within the field.
    \item \textit{Trends: } Current methodologies focus mainly on adapting to fluctuating bandwidth conditions. Given that network capabilities can change significantly in real world settings, particularly within IoT environments, dynamic approaches have been proposed. For example, dynamic gradient compression~\cite{tang2024bandwidth} allows clients to adjust the size of their model updates based on their available bandwidth, allowing clients with superior connections to transmit more detailed updates. Similarly, adaptive model compression techniques, such as model sparsification, have been recently introduced~\cite{liu2025adaptivefl}, which increase compression levels as bandwidth becomes more constrained. This adaptive trend has also been extended to client selection. For example, in~\cite{zhang2023joint}, a deep reinforcement learning agent is trained on the server side to dynamically select clients according to a set of collected network metrics.
    \item \textit{To do: } Future research efforts could explore the joint optimization of multiple dimensions within the training process, such as batch size and model compression, to achieve enhanced performance~\cite{zhang2023joint}. Furthermore, privacy considerations must remain paramount in future endeavors to ensure compliance with the inherent limitations of FL. Finally, establishing robust theoretical foundations is essential for a deeper understanding of the proposed methodologies~\cite{pervej2024hierarchical}.
\end{itemize}

\begin{minorenv}
\begin{enumerate}[label=\textbf{RQ6.2:}, leftmargin=*]
  \item How can FL remain stable and efficient under time-varying bandwidth with theoretical guarantees, while ensuring that adaptation mechanisms do not leak sensitive information?
\end{enumerate} 
\end{minorenv}

\paragraph{\textbf{Challenge 6.3: Energy efficiency and carbon footprint}} 
\major{While offering privacy and data sovereignty benefits, FL can incur significant energy costs due to repeated local training and frequent communication rounds across geographically dispersed devices~\cite{gouissem2024energy}. This overhead not only affects operational costs but also contributes to carbon emissions, especially when deployed at scale or using resource-intensive models such as LLMs.}

\begin{itemize}
    \renewcommand{\labelitemi}{\ding{72}}
    \item \textit{Done:} \major{Efficiency has been extensively investigated within the context of FL due to its inherently distributed nature. As previously discussed, FL was initially conceived with efficiency as a fundamental design goal~\cite{bib:mcmahan16communicationefficient}, acknowledging the heterogeneity of participating devices in practical deployments. The application of FL to IoT and edge computing scenarios has necessitated the development of specialized techniques tailored to energy-constrained devices, which typically rely on battery power~\cite{li2021talk} and operate over wireless networks that impose additional communication costs~\cite{yang2020energy}. Early research efforts primarily focused on model weight or gradient compression strategies optimized for client devices, often intersecting with communication efficiency challenges.}
    
    \item \textit{Trends:} \major{Recent advancements in the field have expanded the scope of optimization to include factors such as packet error rate, which captures transmission errors in wireless communication channels~\cite{de2024improving}. Moreover, emerging approaches incorporate strategies like optimized client selection to minimize overall energy expenditure, as well as hyperparameter tuning aimed at accelerating convergence while simultaneously reducing energy consumption~\cite{marnissi2024quant}.}
    
    \item \textit{To Do:} \major{Ongoing research into emerging communication paradigms, such as 6G wireless networks and blockchain technologies, suggests their potential to enhance energy efficiency within FL systems~\cite{gouissem2024energy}. Furthermore, the absence of comprehensive benchmarks that reflect realistic operational environments has been identified as a significant limitation in current studies~\cite{compaore2025energy}. Finally, continued investigation into advanced data compression techniques for communication is anticipated, indicating that substantial opportunities for improvement remain~\cite{gouissem2024energy}.}
\end{itemize}

\begin{minorenv}
\begin{enumerate}[label=\textbf{RQ6.3:}, leftmargin=*]
  \item How can FL jointly optimize energy/carbon, utility, and trustworthiness constraints, supported by realistic benchmarking and reporting? 
\end{enumerate} 
\end{minorenv}

\subsection{Requirement 7: Accountability}

The seventh requirement of \paco{TAI}, accountability, refers to the need for clear mechanisms that ensure the responsibility, auditability, and verifiability of AI systems throughout their entire lifecycle. \major{Accountability serves as a bridge between ethical principles and their measurable enforcement in deployed AI systems.} Accountability involves the ability to trace decisions, document system behavior, effectively manage risks, and assign liability when adverse outcomes occur. This also includes enabling external audits, maintaining comprehensive records of system development and deployment, and ensuring that users have access to meaningful redress mechanisms. However, in FL, achieving accountability presents unique challenges. The distributed architecture of FL means that data, model updates, and decision logic are fragmented across a network of independent clients, often with limited mutual visibility. This fragmentation complicates efforts to document the provenance of the data, trace how individual contributions affect the global model, and determine responsibility in the event of failures or harmful outputs. In addition, the involvement of multiple stakeholders, from data owners to model developers and platform providers, raises questions about how accountability should be shared or distributed. As FL is increasingly adopted in critical domains such as healthcare, finance, and law enforcement, developing mechanisms for transparent logging, federated auditing, and responsibility attribution is essential to ensure that these systems meet both ethical expectations and regulatory requirements.

\ph{In agentic FL settings, accountability is further complicated by the fact that decision authority may be distributed across multiple agents and orchestration layers, making it insufficient to audit model updates alone; instead, decision provenance across the control plane must also be captured to support responsibility attribution and governance~\cite{SAPKOTA2026103599,zhang2026security}.}

\paragraph{\textbf{Challenge 7.1: Model Auditability}} Model auditability encompasses the comprehensive analytical process of verifying that a given machine learning model consistently fulfills its intended function while adhering to legal frameworks and stakeholder obligations. This examination typically includes data processing methodologies, model update mechanisms, and the impact of these processes on model predictions over time. However, in the context of FL, where transparency regarding data utilization and model updates is inherently challenging to track consistently, comprehensive model auditability becomes largely unfeasible. Consequently, substantial additional efforts are required to ensure that the final FL model remains in compliance with current regulatory mandates.

\begin{itemize}
    \renewcommand{\labelitemi}{\ding{72}}
    \item \textit{Done: } Model auditability has been recognized as a significant challenge since the early stages of FL. Given the distributed nature of FL, which restricts direct access to the data, auditability can alternatively be defined as the ability of a member within the federated schema to verify that other members have fulfilled their assigned roles in the training process~\cite{kairouz2021advances}. Initial research focused on auditing the intermediate steps of the training process. For this, immutable ledger technologies such as blockchain have been explored~\cite{jiang2020federated}, allowing the maintenance of a historical record of the model and facilitating comparisons of its evolution after each step of aggregation and the contribution of an individual client. The use of ZKPs, previously mentioned, has also been investigated for this purpose. 
    \ph{System-level trustworthy distributed learning frameworks have also been proposed outside classical FL, underscoring convergent concerns around end-to-end trust properties, coordination assumptions, and auditability in decentralized learning pipelines~\cite{wang2026tdml}.}
    \ph{Recent studies in medical FL argue that trustworthiness also depends on aligning stakeholder incentives and documenting compliance expectations, for instance through reward-system designs that couple participation with measurable assurance signals~\cite{pandl2025reward}.}

    \item \textit{Trends: } Current research continues to prioritize Blockchain technologies~\cite{kalapaaking2025auditable} as the most viable implementation to achieve model auditability in FL. However, data provenance, as previously discussed, has also emerged as a prominent tool to enable model auditability~\cite{prigent2024enabling}. Despite these developments, the prevailing trend remains focused on enhancing existing methods, particularly those that leverage blockchain.
    \item \textit{To do: } Validating certain aspects of a typical FL training process, such as the correct implementation of security-enhancing mechanisms such as DP, has been claimed to present significant difficulty~\cite{kairouz2021advances}. Furthermore, it has been suggested that quantifying the susceptibility of FL systems to various attacks would allow a clearer and more formal understanding of their robustness. \major{Finally, major challenges in auditability applies also to FL systems such as the lack of standardized metrics or benchmarks for measuring the compliance of a given model~\cite{verma2025can}.}
\end{itemize}

\begin{minorenv}
\begin{enumerate}[label=\textbf{RQ7.1:}, leftmargin=*]
  \item What technical artifacts and standards can make FL auditability operational, verifying compliance without centralizing data and with comparable metrics across deployments?
\end{enumerate} 
\end{minorenv}

\paragraph{\textbf{\ph{Challenge 7.2: Control-Plane Auditability and Decision Provenance}}} 
\ph{Control-plane auditability concerns the ability to verify and reconstruct \emph{why} and \emph{how} FL lifecycle decisions were made in agentic settings, including client selection, objective updates, evaluation gating, and deployment or rollback triggers. In agentic FL, these decisions may be taken autonomously and adaptively, often mediated by LLM-based reasoning or orchestration components. As argued in Section~\ref{sec:Agentic_Dynamic}, failures in the control plane can undermine trust even when learning-plane guarantees remain intact. However, current FL auditability approaches largely focus on model updates and aggregation traces, leaving control-plane decisions under-specified, weakly logged, or non-attributable, thereby limiting accountability.}

\ph{Two important aspects regarding to this challenge: 1) Unlike model auditability, which primarily concerns verifying the provenance and integrity of model updates and training artifacts, control-plane auditability targets the traceability and justification of lifecycle decisions (e.g., client selection, metric changes, evaluation gating, and release/rollback actions) that may be made autonomously and can dominate trust outcomes even when learning-plane guarantees hold. 2) Whereas model auditability reconstructs parameter-level lineage and aggregation traces, control-plane auditability reconstructs the justification, authority, and policy compliance of lifecycle decisions.}

\ph{The implications of these agentic AI and control-plane decisions for system-level trust are formalized in Section~\ref{sec:Agentic_Dynamic}.}

\begin{minorenv}
\begin{itemize}
    \renewcommand{\labelitemi}{\ding{72}}
    \item \textit{Done: } Recent agentic FL visions and workflow proposals highlight the growing role of autonomous coordination and decision-making in FL systems, implicitly motivating the need for control-plane observability and governance~\cite{li2025position,nguyen2025vision,sun2025vision,giusti2025federation}.
    
    \item \textit{Trends: } Trust, risk, and security management perspectives for agentic systems increasingly emphasize governance, traceability, and lifecycle assurance as essential complements to technical safeguards~\cite{raza2025trism,zhang2026security}. Parallel work on large-scale agent coordination highlights verifiable collaboration as a system requirement, suggesting that provenance should extend beyond model artifacts to decision artifacts~\cite{sarker2025advancing}.
    \item \textit{To do: } Operational auditability for the control plane requires standardizing decision logs, policy/versioning semantics, and provenance schemas that can attribute outcomes to autonomy-level decisions without leaking sensitive local information. A major open problem is to define interoperable evidence interfaces (e.g., for objective changes, evaluation gating, and release decisions) that are comparable across deployments and can be audited post hoc.
\end{itemize}
\end{minorenv}

\begin{minorenv}
\begin{enumerate}[label=\textbf{RQ7.2:}, leftmargin=*]
  \item What minimal set of control-plane artifacts (decision logs, policy versions, evaluation provenance) makes agentic FL auditable and accountable without centralizing data or revealing proprietary coordination policies?
\end{enumerate}
\end{minorenv}

\paragraph{\textbf{Challenge 7.3: Legal Responsibility}} Despite concerted recent regulatory efforts, the prevailing legal landscape surrounding FL remains ambiguous, with existing statutes lacking explicit clarity on specific terms. A notable illustration of this challenge is elucidated in \cite{herbert2024federated}, which highlights a problematic intersection with the AI Act of the European Union. The AI Act mandates clear delineation of stakeholder responsibilities throughout the development and deployment lifecycle of a machine learning model. However, within the FL paradigm, both the server and the participating clients inherently share responsibilities from a legal point of view. This shared accountability necessitates further clarification regarding the FL paradigm and introduces open regulatory challenges that must be addressed to improve the feasibility and broader adoption of FL under the provisions of the AI Act. Currently, there are no significant efforts to tackle this challenge, making it a problem that must be addressed.

\begin{itemize}
    \renewcommand{\labelitemi}{\ding{72}}
    \item \textit{To do: } The allocation of legal responsibility within FL, particularly with respect to stakeholder obligations, has recently been identified as a significant challenge~\cite{herbert2024federated}. \major{Moreover, a cross-border FL deployment may introduce additional complexity in determining applicable jurisdiction and data protection obligations. For instance, when clients are located in multiple legal regions, questions arise regarding which data protection laws prevail and how liability is shared under joint controllership models, as defined by the GDPR. However, such contractual models remain largely theoretical and lack legal precedent}. This issue requires further investigation and the development of a more refined legal framework to provide much-needed clarification.
\end{itemize}

\begin{minorenv}
\begin{enumerate}[label=\textbf{RQ7.3:}, leftmargin=*]
  \item What governance frameworks and technical evidence can delineate stakeholder responsibilities and jurisdiction in cross-silo FL under evolving regulatory regimes? 
\end{enumerate} 
\end{minorenv}

\paragraph{\textbf{Challenge 7.4: Data Provenance}}
As described in Challenge~4.4, data provenance involves meticulously documenting the origin and complete lifecycle of data used to train a machine learning model, encompassing details of its collection and subsequent transformations~\cite{souza2019provenance}. This practice empowers users, auditors, and stakeholders to verify the trustworthiness and reliability of the data underpinning machine learning systems, facilitating a more effective tracing of biases, inconsistencies, or model errors, thus becoming a crucial component of accountability. However, as previously noted, maintaining such comprehensive records within FL environments poses substantial challenges due to the inherent privacy-preserving nature of the paradigm and the impossibility (and undesirability) of centralized raw-data retention~\cite{gu2024enhancing}. \major{Moreover, the broader AI landscape shows that the authenticity, consent, and provenance infrastructures remain fragmented and “broken,” underscoring the need for interoperable, verifiable provenance standards and libraries that FL can adopt in a privacy-thrifty form~\cite{longpre2024data}. \ph{Blockchain-backed logging has been explored as a practical substrate for audit trails and tamper-evident coordination metadata in trustworthy edge FL deployments, reinforcing the importance of persistent provenance evidence beyond model weights alone~\cite{Moore2025}.} As a result, the “work done / trends / to-do” items here overlap with Challenge~4.4: provenance is simultaneously a transparency and an accountability enabler in FL.}

\begin{minorenv}
\begin{enumerate}[label=\textbf{RQ7.4:}, leftmargin=*]
  \item How can interoperable provenance mechanisms support incident response and liability attribution in FL while remaining privacy-thrifty and robust to adversarial tampering? 
\end{enumerate} 
\end{minorenv}

\subsection{Summary Challenge Taxonomy and Blueprint for TFL}\label{sec3:summary_table}

In Figure~\ref{fig:key_challenges} and Table \ref{tab:key_gaps_short}  we summarize the main key gaps / remaining work found when following TFL. They are organized according to the seven TAI requirements, highlighting the main challenges taxonomy and the blueprint for each one. 

\begin{figure}[htbp!]
    \centering
  \includegraphics[width=0.5\textwidth]{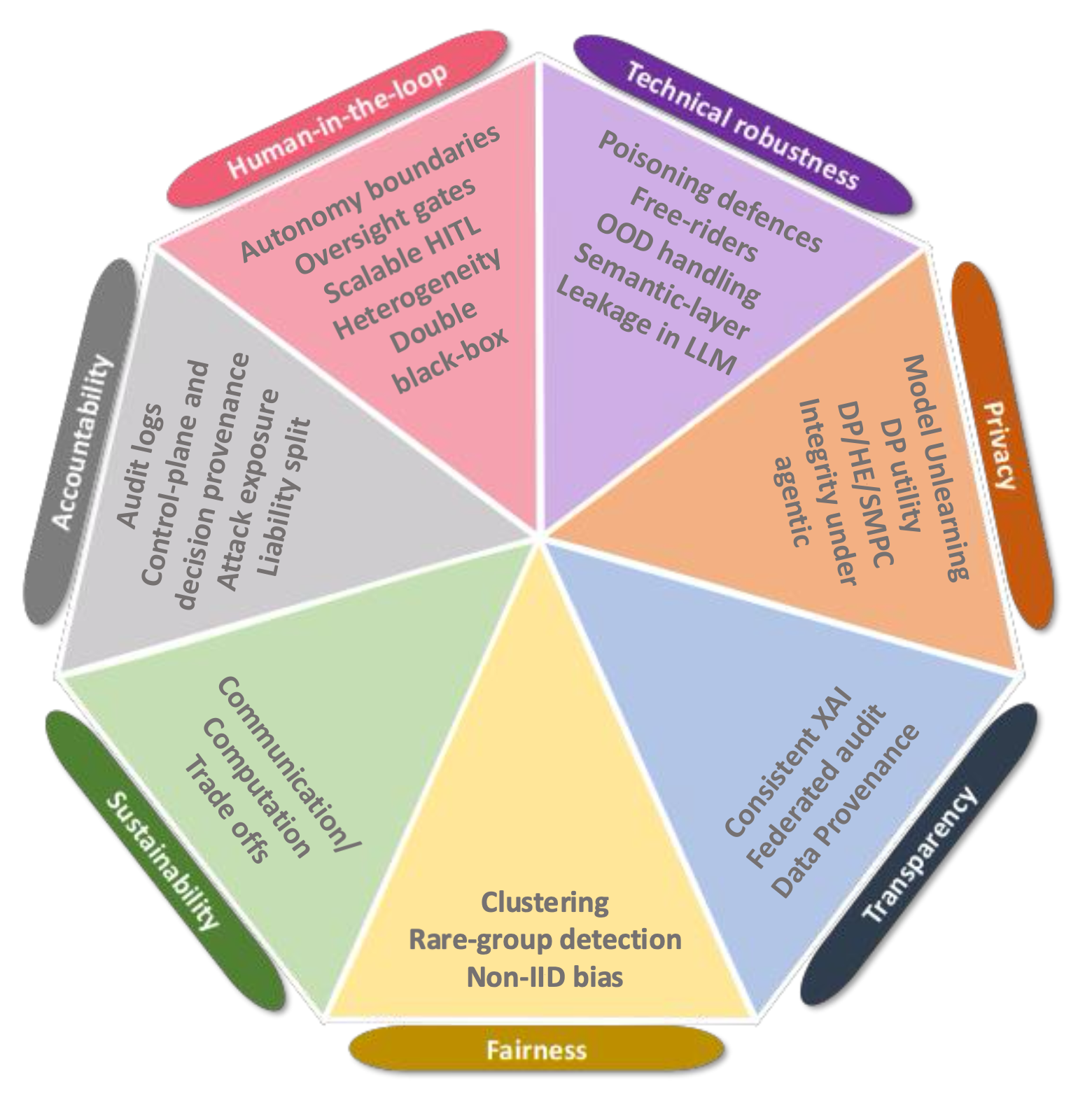}
  \caption{Visual representation of the main key challenges  in TFL. For the sake of clarity, we use the same color theme that in Figure~\ref{fig:example}.}
  \label{fig:key_challenges}
\end{figure}

\begin{longtable}{p{0.2\linewidth} p{0.75\linewidth}}
\caption{Challenges taxonomy and blueprint of TFL organized by TAI requirements.}
\label{tab:key_gaps_short} \\
\toprule
\textbf{Dimension} & \textbf{ Challenges taxonomy and blueprint} \\
\midrule
\endfirsthead

\multicolumn{2}{c}{{\tablename\ \thetable{} continued from previous page}} \\
\toprule
\textbf{Dimension} & \textbf{Main Challenges} \\
\midrule
\endhead

\midrule 
\multicolumn{2}{r}{} \\
\endfoot

\bottomrule
\endlastfoot

\textbf{Human-in-the-loop} &
\textbf{Scalable HITL:} \major{Designing HITL mechanisms that can scale across thousands of clients in federated settings.}
 
 \textbf{Heterogeneity:} \major{Managing diverse data and system conditions that complicate consistent human feedback.}

\textbf{Double black-box:} \major{Enhancing interpretability of both client-level and aggregated decisions, especially in Vertical FL.}

\textbf{\ph{Autonomy boundaries and oversight gates:} }\major{Constraining agentic clients/coordinators with explicit autonomy limits and runtime oversight gates to prevent unsafe or misaligned actions during federated coordination.}

\\[1em]

\textbf{Privacy and Data Governance} &
\textbf{DP utility trade-off:} \major{Preserving privacy through DP while minimizing performance degradation.}
 
  \textbf{Hybrid privacy stacks:} \major{Combining DP, HE and SMPC efficiently.}

\textbf{Model unlearning:} \major{Enabling the removal of a client’s contribution without full retraining, in line with GDPR requirements.}

\textbf{\ph{Evaluation integrity under agentic coordination:}} \major{ Preventing agents from manipulating validation protocols, metrics, or reporting to inflate performance or hide failures, preserving trustworthy evaluation in TFL.}

 \\[1em]

\textbf{Technical Robustness and Safety} &
\textbf{Poisoning defence:} \major{Developing lightweight yet strong defences against data and model update poisoning.}

 \textbf{Free rider detection:} \major{Identifying clients that benefit without contributing valid updates.}
 
 \textbf{OOD:} \major{Building models that reliably detect and manage out-of-distribution data in non-IID settings.}

 \textbf{\ph{Semantic-layer leakage in LLM-enabled FL:} }\major{Addressing privacy leakage at the semantic/generative level (memorization, prompts, outputs) and orchestration-induced disclosure, beyond gradient-level protections such as DP.}

\\[1em]

\textbf{Transparency and Explainability} &
 \textbf{Consistent XAI:} \major{Creating federated explainability frameworks consistent across heterogeneous clients.}

\textbf{Federated audit:} \major{Establishing mechanisms to trace, align, and version model explanations for accountability.}

 \textbf{Data provenance:} \major{Tracking contributions and transformations without revealing raw data (e.g., via watermarking or ZKPs).} \\[1em]

\textbf{Fairness and Diversity} &
 \textbf{Non-IID bias:} \major{Mitigating bias from uneven or skewed data distributions across clients.}

 \textbf{Privacy safe clustering:} \major{Grouping clients without leaking sensitive distribution information.}

\textbf{Rare group detection:} \major{Ensuring minority or underrepresented client populations are adequately represented.} \\[1em]

\textbf{Sustainability and Societal Well-being} &

 \textbf{Carbon efficiency:} \major{Optimizing client selection and compression to minimize environmental impact.}
  
   \textbf{Benchmarking:} \major{Developing standardized metrics to compare sustainability trade-offs across FL systems.} \\[1em]

\textbf{Accountability} &
\textbf{Audit logs:} \major{Implementing verifiable federated audit trails, possibly blockchain-based.}

\textbf{\ph{Control-plane auditability and decision provenance:}} \major{Ensuring end-to-end traceability of orchestration decisions in agentic/LLM-driven TFL to enable audits, accountability, and compliance.}

 \textbf{Attack exposure quantification:} \major{Measuring vulnerability of FL systems to various threats.}

 \textbf{Liability split:} \major{Defining legal responsibility among servers, clients, and stakeholders.} \\[1em]

\end{longtable}

\section{\ph{Trustworthiness in Agentic AI and Dynamic Federated Learning}}
\label{sec:Agentic_Dynamic}

\ph{FL systems are increasingly embedded within agentic AI architectures, where autonomous components make decisions beyond local model training, and where LLMs act as reasoning, coordination, or control elements. We further consider dynamic environments as settings characterized by temporal non-stationarity, including concept drift, evolving client populations, and changing operational constraints, which differ fundamentally from static non-IID data distributions. Under these conditions, trustworthiness must be understood as a system-level and lifecycle-dependent property, rather than a one-shot guarantee established at training time.}

\paragraph{\ph{Definitions and scope}}
\ph{In this paper, we use \emph{agentic FL} to denote FL systems in which autonomous components can make and adapt control-plane decisions across the FL lifecycle (e.g., adapt objectives, select participants, configure training workflows, or trigger deployment and retraining actions with limited human intervention), rather than merely executing a fixed training pipeline \cite{acharya2025agentic,jiang2025large,nguyen2025vision}. 
We use \emph{dynamic environments} to denote temporal non-stationarity coupled with adaptive control, including concept drift, evolving client populations, and changing operational constraints that may trigger autonomous reconfiguration. Accordingly, trustworthiness is treated as a continuously maintained, lifecycle-dependent system property, not a one-shot guarantee established at training time \cite{polato2026learning}.}

\ph{This section is intentionally conceptual: its goal is not to propose new mechanisms, but to formalize how agency and dynamics reshape the trust problem itself, thereby motivating the coordination and assurance constructs introduced later. Where prior FL surveys primarily catalog technical vulnerabilities or optimization challenges, this section reframes trustworthiness itself as a function of agency, autonomy, and lifecycle governance in federated systems.}

\subsection{\ph{From Learning Pipelines to Agentic Control Planes}}
\label{subsec:Learning_vs_Control}

\ph{Traditional FL architectures are commonly modeled as learning pipelines, in which trustworthiness is primarily assessed through properties of the training process itself, such as privacy preservation, robustness to adversarial updates, and convergence guarantees. Within this paradigm, trust is typically evaluated at training time and is closely tied to model-centric assurances, including secure aggregation, differential privacy, and robustness defenses. While these mechanisms remain necessary, they implicitly assume that the learning objective, evaluation criteria, and deployment context are fixed and externally governed.}

\ph{In contrast, agentic FL systems introduce a distinct \emph{control plane} that governs decision-making processes beyond parameter optimization, including client selection, objective updates, evaluation gating, scheduling, and deployment or rollback decisions~\cite{li2025position,sun2025vision,giusti2025federation}. These decisions are often adaptive and may be taken autonomously in response to system feedback or environmental changes. As a result, trustworthiness can no longer be attributed solely to the correctness of the learning algorithm, but must also account for how and why such control decisions are made over time.}

\ph{When LLMs are integrated into federated systems, the role of the control plane is further expanded. LLMs may act as reasoning engines, coordination modules, or semantic interfaces that interpret system state, generate policies, or orchestrate complex workflows across distributed participants~\cite{amini2025distributed,wei2025federated,kristiani2026deploying}. While these capabilities enable more flexible and scalable coordination, they also introduce additional opacity and potential misalignment, as control decisions may depend on learned representations, prompts, or tool invocation chains that are difficult to anticipate or audit.}

\ph{The explicit separation between the learning plane and the control plane is therefore essential for reasoning about trustworthiness in agentic FL systems. Failures or misalignments in the control plane, such as inappropriate client selection, metric manipulation, or premature deployment, can undermine trust even when the underlying learning process remains technically sound. By distinguishing these two planes, it becomes possible to analyze trust not only in terms of model behavior, but also in terms of governance, oversight, and accountability for autonomous system-level decisions.}

\ph{Importantly, the learning and control planes are analytically orthogonal: a system may satisfy strong learning-time guarantees while remaining untrustworthy due to failures in control-plane decision-making. }

\ph{Figure~\ref{fig:learning_control_plane} illustrates the conceptual separation between the learning plane and the control plane in agentic FL systems. While the learning plane governs model-centric operations such as local training and aggregation, the control plane encapsulates autonomous decision-making related to coordination, evaluation, and lifecycle management. This separation is essential for reasoning about trustworthiness, as failures in control-plane decisions may undermine system trust even when learning-plane guarantees remain intact.}

\begin{figure}[th!]
\centering
\includegraphics[width=0.7\linewidth]{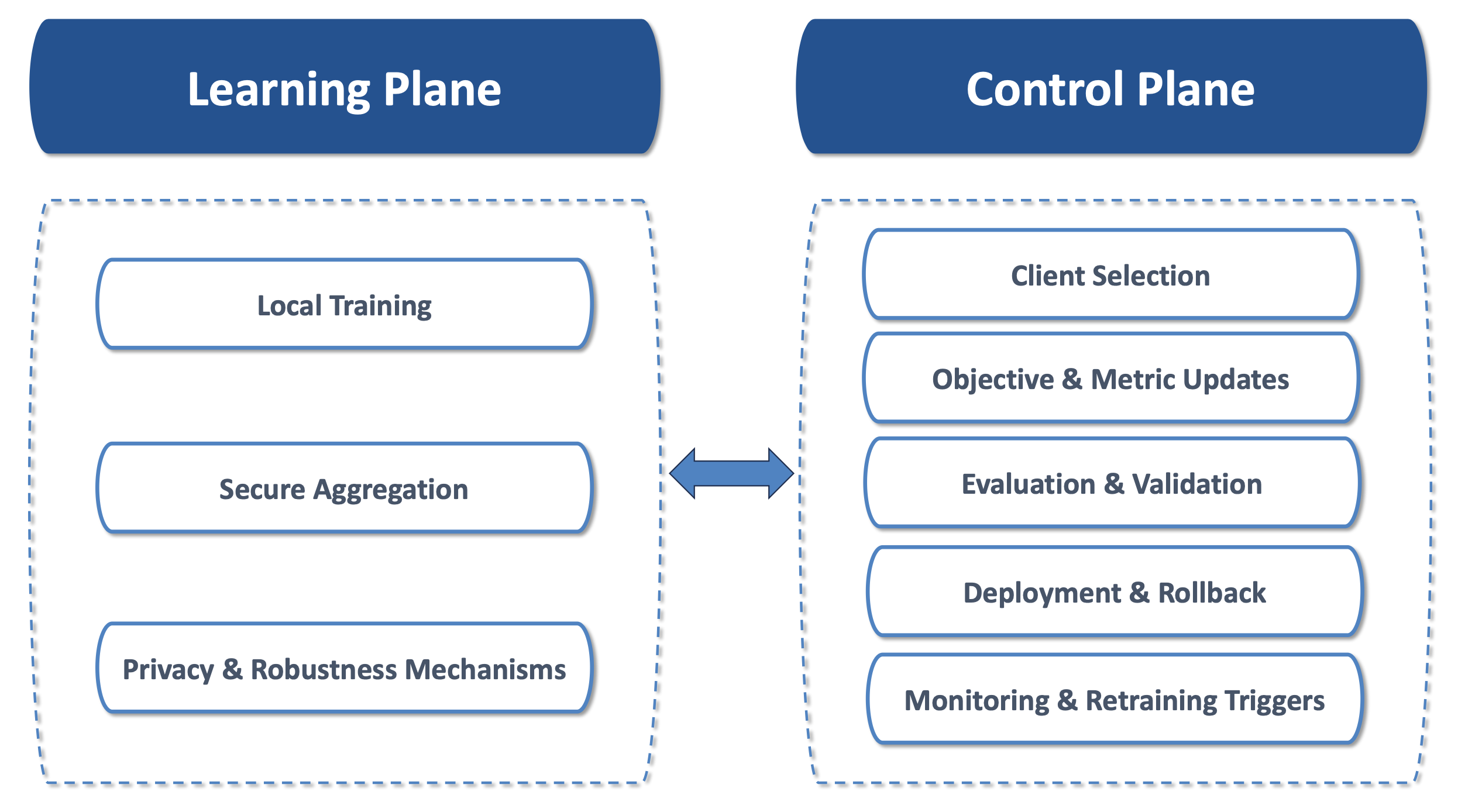}
\caption{\ph{Conceptual separation between the learning plane and the control plane in agentic FL systems. Feedback and signals exchanged between the two planes enable adaptive control, continuous monitoring, and the generation of auditable evidence that underpins trustworthiness over time.} }
\label{fig:learning_control_plane}
\end{figure}

\subsection{\ph{Agency Decomposition Across the Federated Lifecycle}}
\label{subsec:Agency_Decomposition}

\ph{To make agency analyzable and governable in FL, it is necessary to explicitly decompose where and how autonomous decisions are exercised across the federated lifecycle. In classical FL settings, most design choices, such as objectives, evaluation criteria, and deployment policies, are typically fixed ex ante and managed outside the learning system. In contrast, agentic FL systems allow these decisions to be adapted dynamically, often in response to observed performance, environmental feedback, or system-level constraints.}

\ph{In such settings, agency may be distributed across multiple decision loci spanning the entire lifecycle, including the selection of optimization objectives and evaluation metrics, participant sampling and admission control, training configurations (e.g., learning rates, privacy budgets, or adapter strategies), aggregation rules, evaluation design, and deployment or retraining triggers~\cite{nguyen2025vision,sarker2025advancing}. Each of these loci represents a point at which autonomous decisions can shape system behavior and outcomes, and therefore a potential source of both capability and risk from a trustworthiness perspective.}

\ph{When LLMs are integrated as reasoning or coordination components, the scope of agency is further expanded. In these cases, autonomous decisions may arise from prompt formulation, semantic interpretation of system state, tool invocation, or the generation of high-level policies that guide downstream actions~\cite{sun2025generalizing,tang2026rethinking}. Because such decisions are mediated by learned representations and language-based abstractions, they may be less transparent and more difficult to attribute to specific design choices, increasing the importance of explicit agency modeling and documentation.}

\ph{This decomposition of agency across decision loci enables a more precise analysis of how autonomous actions affect different trust dimensions, such as fairness, accountability, robustness, and transparency. For example, decisions related to participant selection or evaluation design may disproportionately impact fairness, while deployment or retraining triggers raise questions of accountability and traceability. By identifying where agency is exercised, this perspective provides a foundation for targeted oversight, principled autonomy boundaries, and systematic evidence collection, which are essential for maintaining trustworthiness throughout the FL lifecycle, as summarized across autonomy levels in Table~\ref{tab:Decision_Autonomy_Matrix}. This agency decomposition is later operationalized through control-plane audit artifacts in Section \ref{Sec:TAI-Blueprint}.}

\subsection{\ph{Levels of Autonomy and Human Oversight}}
\label{subsec:Autonomy_Levels}

\ph{Agentic FL systems exhibit varying degrees of autonomy, which have direct and non-linear implications for trustworthiness, accountability, and governance. In contrast to classical FL deployments, where most decisions are externally managed and human-controlled, agentic systems increasingly delegate decision-making authority to autonomous components. As a result, the degree of autonomy exercised by agents becomes a critical design parameter that shapes both system behavior and the nature of trust that can be reasonably established.}

\ph{We distinguish between multiple levels of autonomy, ranging from fully manual operation, in which human operators retain direct control over all decisions, to semi-autonomous configurations where agents provide recommendations or execute actions within predefined policies, and fully autonomous settings in which agents can modify objectives, workflows, or coordination strategies end-to-end~\cite{acharya2025agentic,raza2025trism}. These levels reflect not only technical capabilities but also governance choices regarding which decisions may be delegated, under what conditions, and with what constraints. Importantly, autonomy should not be viewed as a binary property, but rather as a spectrum that must be explicitly characterized and managed.}

\ph{As autonomy increases, the potential impact of individual decisions on system outcomes and stakeholders also grows, amplifying the need for explicit guardrails and human oversight. In agentic FL, high-impact decisions, such as model release or rollback, changes to optimization objectives, or adjustments to privacy budgets, can have far-reaching consequences for safety, fairness, and compliance. Without clearly defined escalation mechanisms and approval gates, fully autonomous actions may lead to irreversible trust failures, even in the absence of malicious behavior.}

\ph{Clear articulation of autonomy levels and their associated oversight requirements is therefore a prerequisite for maintaining trust in agentic FL systems. By explicitly specifying which decisions are automated, which require human validation, and which must remain under direct human control, system designers can align autonomy with TAI requirements and establish accountability boundaries. This perspective also provides a natural basis for auditability, as autonomy levels determine what evidence must be recorded and reviewed to justify system behavior over time.}

\paragraph{\ph{Decision loci and autonomy levels}}
\ph{To make agency and autonomy operational rather than purely conceptual, it is necessary to explicitly relate where decisions are made in the FL lifecycle to the degree of autonomy with which they may be exercised. Different decision loci entail different risk profiles, oversight requirements, and trust implications depending on whether decisions are made manually, recommended by agents, executed under predefined policies, or delegated end-to-end. Table~\ref{tab:Decision_Autonomy_Matrix} summarizes this relationship by mapping key decision loci in agentic FL systems to increasing levels of autonomy, thereby clarifying which decisions require strict human oversight and which may be safely automated under appropriate governance constraints.}

\begin{minorenv}
    
\begin{table}[ht!]
\centering
\begin{colortabular}{black}
\begin{tabular}{p{2.7cm} p{2.2cm} p{2.2cm} p{2.2cm} p{2.2cm}}
\toprule
\textbf{Decision Locus} &
\textbf{A0: Manual} &
\textbf{A1: Agent Recommends} &
\textbf{A2: Agent Executes (Policy)} &
\textbf{A3: Fully Autonomous} \\
\midrule
Objective and metric selection &
Human-defined &
Suggested by agent &
Auto-selected within policy &
Agent-defined and modified \\[0.3em]

Client participation and sampling &
Human-approved &
Agent proposes cohort &
Automated under constraints &
Autonomous admission/exclusion \\[0.3em]

Training configuration (e.g., epochs, DP budget, adapters) &
Human-configured &
Agent suggests parameters &
Automated within bounds &
Agent adapts configuration \\[0.3em]

Aggregation rule selection &
Human-selected &
Agent recommends rule &
Policy-constrained execution &
Autonomous strategy switching \\[0.3em]

Evaluation design and testing &
Manually defined &
Agent suggests tests &
Automated evaluation pipeline &
Adaptive evaluation criteria \\[0.3em]

Deployment and rollback decisions &
Human-controlled &
Agent alerts and advises &
Automated under approval rules &
Autonomous deployment/rollback \\[0.3em]

Monitoring and retraining triggers &
Manual monitoring &
Agent flags anomalies &
Automated retraining triggers &
Self-initiated adaptation \\
\bottomrule
\end{tabular}
\end{colortabular}
\caption{\ph{Decision loci across the FL lifecycle and corresponding autonomy levels.}}
\label{tab:Decision_Autonomy_Matrix}
\end{table}
\end{minorenv}

\subsection{\ph{Agency-Specific Threat Models in Dynamic Environments}}
\label{subsec:Agency_Threats}

\ph{Many agency-specific threats emerge at higher autonomy levels (A2–A3), particularly for deployment and evaluation decisions.}

\ph{Classical FL threat models primarily focus on adversarial behaviors at the level of model updates or data, such as poisoning attacks, inference attacks, or Byzantine participation. While these threats remain relevant, they implicitly assume that the learning objective, evaluation criteria, and system workflow are fixed and externally governed. In agentic and dynamic FL systems, this assumption no longer holds, as autonomous components may adapt objectives, evaluation strategies, or coordination policies over time, introducing new avenues for trust failures that are not captured by traditional threat models.}

\ph{In such settings, agency gives rise to qualitatively different classes of risks that directly affect trustworthiness. These include Goodhart-style metric gaming, where agents optimize proxy objectives or performance indicators at the expense of underlying system goals; manipulation of evaluation procedures through selective testing, adaptive benchmark selection, or suppression of regression signals; and abuse of orchestration mechanisms or toolchains in LLM-enabled workflows, where agents may modify prompts, invoke tools opportunistically, or reconfigure pipelines in unintended ways~\cite{sun2025generalizing,tang2026rethinking,zhang2026security}. Unlike classical attacks, these behaviors may emerge from misaligned incentives or adaptive optimization rather than explicit adversarial intent.}

\ph{Dynamic environments further exacerbate these risks through the interaction between autonomy and temporal non-stationarity. Concept drift, evolving client populations, or changing operational constraints can interact with autonomous retraining or policy adaptation, leading to feedback loops in which corrective actions amplify instability, degrade fairness across subpopulations, or mask gradual performance regressions. In such cases, trust failures may accumulate silently over time, making them difficult to detect through static validation or one-shot evaluation procedures.}

\ph{In regulated and high-risk domains, such as healthcare IoT and clinical FL, agency-specific threats are compounded by strict safety, accountability, and compliance requirements~\cite{adil2026security,hanna2026infectious}. Here, failures in evaluation integrity, deployment control, or drift management can have direct real-world consequences and legal implications. Addressing these risks therefore requires moving beyond static threat models toward lifecycle-aware risk assessment, continuous monitoring, and auditable evidence of decision-making processes, capabilities that are explicitly operationalized through the coordination blueprint and Trust Report~2.0 introduced in the following sections.}

\paragraph{\ph{Mapping agency and dynamics to Trustworthy AI requirements}}
\ph{While the preceding analysis focuses on agency, autonomy, and dynamics as system-level properties, these concepts must ultimately be grounded in the TAI requirements introduced in Section~\ref{sec3:challenges}. Rather than constituting an independent conceptual layer, agency-aware FL  reshapes how each TAI requirement is instantiated, monitored, and governed over time. Table~\ref{tab:Agentic_TAI_Mapping} summarizes how the key dimensions analyzed in this section, control-plane decision-making, distributed agency, autonomy levels, and dynamic threat models, map onto the seven TAI requirements, thereby ensuring conceptual continuity between the taxonomy, the coordination blueprint, and the Trust Report~2.0 framework.}

\begin{table}[ht!]
\centering
\begin{colortabular}{black}
\begin{tabular}{p{3.2cm} p{9.5cm}}
\toprule
\textbf{TAI Requirement} & \textbf{Implications in Agentic and Dynamic FL} \\
\midrule
Human oversight (R1) &
Autonomous control-plane decisions require explicit autonomy boundaries, escalation mechanisms, and approval gates for high-impact actions such as model release, rollback, or objective changes. \\[0.3em]

Robustness and safety (R2) &
Dynamic environments and adaptive retraining introduce feedback loops and instability risks, requiring continuous monitoring and safeguards beyond static robustness guarantees. \\[0.3em]

Privacy and data governance (R3) &
Agency expands the privacy attack surface through adaptive client selection, semantic communication, and policy updates, necessitating lifecycle-aware privacy controls and boundary enforcement. \\[0.3em]

Transparency and traceability (R4) &
Distributed agency and LLM-enabled orchestration increase opacity, making decision provenance, logging, and traceability across the control plane essential for understanding system behavior. \\[0.3em]

Fairness and non-discrimination (R5) &
Autonomous decisions related to participant sampling, evaluation design, or retraining triggers can induce or amplify subgroup-level disparities under drift, requiring fairness assessment over time. \\[0.3em]

Societal and environmental well-being (R6) &
Continual adaptation and repeated retraining raise concerns about computational cost, energy consumption, and sustainability, particularly in large-scale or edge-deployed federated systems. \\[0.3em]

Accountability (R7) &
Agency decomposition and autonomy levels determine responsibility attribution, as trust failures may stem from control-plane decisions rather than learning outcomes, requiring auditable evidence for post-hoc analysis. \\
\bottomrule
\end{tabular}
\end{colortabular}
\caption{\ph{Mapping of agentic and dynamic FL dimensions to TAI  requirements.}}
\label{tab:Agentic_TAI_Mapping}
\end{table}

\paragraph{\ph{Trustworthiness in agentic and dynamic FL}}
\ph{Taken together, these considerations answer Q3 by showing that trustworthiness in agentic and dynamic FL cannot be ensured solely through model-centric guarantees or static design-time assumptions as summarized in Table~\ref{tab:Agentic_TAI_Mapping}. Instead, autonomous decision-making, evolving objectives, and environmental non-stationarity require trust to be continuously established and maintained through explicit control of agency, well-defined autonomy boundaries, and systematic monitoring of decision-making processes across the federated lifecycle. These observations motivate the coordination blueprint introduced in the next section, which structures interactions between the learning and control planes and surfaces cross-requirement trade-offs under autonomy. Moreover, they justify the need for a lifecycle-oriented Trust Report, extended in this work as Trust Report~2.0, which operationalizes trustworthiness by capturing auditable signals related to agency, dynamics, and governance over time, thereby translating the conceptual insights of this section into actionable and verifiable system-level assurances.}

\section{\major{A Coordination Blueprint for Trustworthy Federated Learning Systems: Through the Lens of Trust Report 2.0 }}\label{Sec:TAI-Blueprint}

\ph{The challenges identified in Section~\ref{sec3:challenges} and the agentic, dynamic considerations discussed in Section~\ref{sec:Agentic_Dynamic} reveal that TFL cannot be achieved by satisfying individual TAI requirements in isolation. Instead, trustworthiness emerges from how these requirements interact, trade off, and are jointly governed across the FL lifecycle. This section therefore proposes a \emph{coordination blueprint} whose purpose is not to introduce new learning mechanisms, but to structure how technical, organizational, and governance decisions are aligned when multiple trust dimensions are simultaneously at stake. By making requirement interactions explicit and associating them with decision points, oversight roles, and auditable evidence, the blueprint provides a system-level foundation for explainable, accountable, and governable TFL.}

\ph{In this blueprint, trustworthiness is operationalized by governing decisions rather than models, explicitly connecting TAI requirements to federated controls and the evidence required to justify them throughout the lifecycle.}

\paragraph{\ph{Coordination as a lifecycle process}}
\ph{At a high level, the coordination blueprint can be understood as a recurring lifecycle process rather than a one-off configuration activity. In TFL, trust-relevant issues arise continuously during training, deployment, and adaptation due to factors such as data drift, changing participation, evolving objectives, and autonomous control-plane behavior. The blueprint therefore structures coordination as a loop in which issues are first detected, their implications across multiple TAI requirements are assessed, decisions are taken under explicit governance and oversight constraints, and the resulting actions are recorded as auditable evidence. As illustrated in Figure~\ref{fig:Loop-TFL}, this recurring coordination loop ensures that trustworthiness is maintained over time through documented decisions and proportional evidence, rather than assumed from static design-time guarantees.}

\begin{figure}[th!]
\centering
\includegraphics[width=0.7\linewidth]{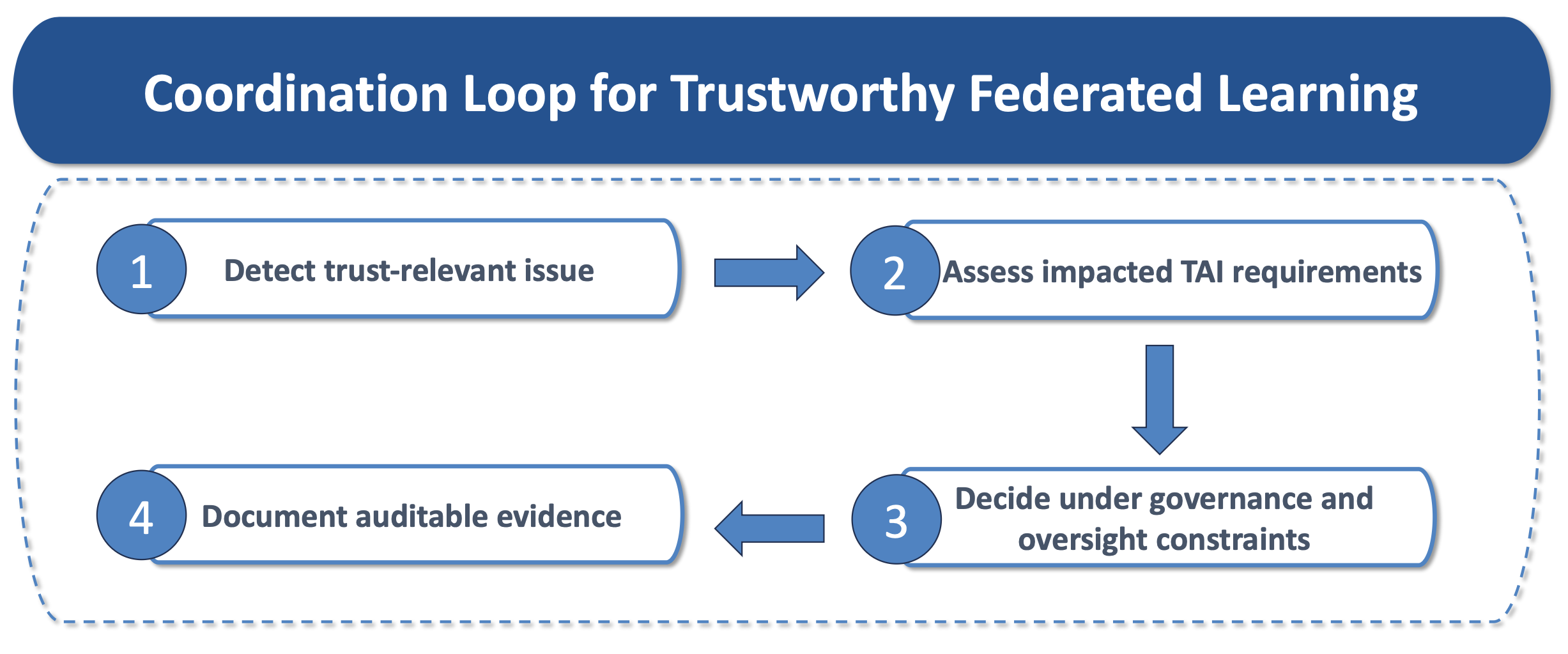}
\caption{\ph{Recurring coordination loop for TFL.}}
\label{fig:Loop-TFL}
\end{figure}

\subsection{\ph{Blueprint for Trustworthy Federated Learning Systems}}

\major{While the \rewrite{previous} sections discuss each challenge in isolation, in practice many are deeply interdependent. Improving one dimension of TFL \rewrite{may impact} another, giving rise to fundamental trade-offs that must be \rewrite{considered} through technical design and governance. Understanding these interdependencies is essential for aligning FL with the full set of TAI requirements.}

\rewrite{Rather than treating these interdependencies as obstacles, we view them as explicit design choices managed through FL-native controls. As recent literature suggests, privacy is a system-wide property that coordinates with robustness, fairness, and accountability \cite{fang2025trustworthy, saeed2024security}. For instance, privacy risks often propagate through model updates and intersect with adversarial vulnerabilities, requiring that transparency and fairness be engineered jointly rather than traded off naively \cite{wei2025trustworthy, yang2024robustness}. Consequently, our approach prioritizes design elements such as local-first processing, privacy-aware aggregation, and interpretable reporting to ensure that these trade-offs are negotiated, measurably safe, and compliant \cite{radanliev2025ai, nadella2025privacy}.}

We operationalize Trust beyond privacy by explicitly linking: \\

\begin{center}
Principles $\rightarrow$ FL-native Controls $\rightarrow$ Accountable Evidence.
\end{center}

\paragraph{\major{What we mean by FL-native controls}}
We define FL-native controls as technical and governance mechanisms that operate {on the edge and in the round}, \rewrite{that is}, they assume data stays local, updates are coordinated asynchronously across heterogeneous clients, and only privacy preserving information leave the node. Unlike generic AI controls, FL-native controls bind each TAI requirement to actions that are enforceable in federated settings and that yield evidence \rewrite{(artifacts)} \rewrite{that foster accoutability} without exposing individuals.

\rewrite{As such, we enumerate the natural trade-offs that arise between pillars in TFL and give a Blueprint as actionable advice of how to coordinate them through the use of FL-native controls.}

\paragraph{\ph{Decision-centric coordination}}
\ph{Rather than optimizing individual requirements in isolation, the blueprint organizes coordination around recurring decision types that simultaneously affect multiple trust dimensions, require different levels of oversight, and produce distinct evidence artifacts.}

\ph{Table~\ref{tab:Blueprint_Decisions} summarizes how recurring decision types in FL simultaneously affect multiple TAI requirements, the corresponding evidence needed for justification, and the level of oversight required.}

\begin{table}[ht!]
\centering
\begin{colortabular}{black}
\begin{tabular}{p{3.2cm} p{3.1cm} p{3.8cm} p{2cm}}
\toprule
\textbf{Decision Type} &
\textbf{Affected TAI Requirements} &
\textbf{Required Evidence} &
\textbf{Oversight Level} \\
\midrule
Client admission or exclusion &
R1, R3, R5 &
Decision rationale, policy version, cohort statistics &
Human-in-the-loop \\[0.3em]

Objective or metric change &
R2, R5, R7 &
Change justification, impact analysis, prior metrics &
Human approval \\[0.3em]

Model release or rollback &
R1, R2, R7 &
Evaluation results, regression checks, approval log &
Human approval \\[0.3em]

Automated retraining trigger &
R2, R5, R6 &
Drift indicators, trigger thresholds, retraining logs &
Policy-constrained \\[0.3em]

Privacy budget update &
R3, R7 &
Budget usage, risk assessment, authorization record &
Human approval \\
\bottomrule
\end{tabular}
\end{colortabular}

\caption{\ph{Decision-centric view of coordination in TFL}}
\label{tab:Blueprint_Decisions}
\end{table}

\ph{Several decision classes in Table~\ref{tab:Blueprint_Decisions} already appear in deployed or proposed TFL systems, such as incentive and participation policies that influence collaboration quality and fairness~\cite{rashid2025trustworthy,pandl2025reward}.}

\paragraph{\textbf{Fairness vs. Efficiency}}
\major{Fairness measures, \rewrite{such as} personalization \rewrite{or} balanced aggregation, may slow convergence \rewrite{and introduce additional costs}. \rewrite{Nevertheless, FL-native controls like} coordinated client sampling, cost-aware scheduling, and \rewrite{model compression} \rewrite{can} recover much of the efficiency \rewrite{via minimizing communitacion costs} without sacrificing equity\rewrite{~\cite{bharati2022federated}}.} \\
\rewrite{Coordination of this pillars can be carried out by jointly planning the participation quotas and the communication budge of each client where applicable. This could latter be reflected as an evidence via participation distributions for each client anonymized with privacy preserving techniques such as DP in order to not leak sensitive information.}

\paragraph{\textbf{Privacy vs. Robustness}}
\major{Privacy controls \rewrite{like DP} can blur signals used for attack detection \rewrite{making it harder for methods like robust aggregation techniques to tell apart benign updates from malicious ones}, yet \rewrite{carefully calibrated} clipping \rewrite{and} noise schedules and client reputation \rewrite{techniques like the ones found in Blockchain architectures~\cite{rehman2020towards} are FL-native controls that allow to} achieve robust performance with bounded leakage.} \\
\rewrite{While the coordination of these requirements may be tricky, we suggest to use red teaming techniques such as synthetic experiments that will help to align the privacy budgets with robust aggregation allowing for sensitive choices for both tasks. This could later be reflected into the findings of this red teaming techniques and more precise metrics of the FL environment such as the privacy budget per round, enabling stakeholders to easily verify the choices made and their outcome.}

\paragraph{\textbf{Transparency \& explainability vs. Privacy}}
\rewrite{As previously stated, an excessive disclosure of information of an individual for the sake of transparency may lead to potential privacy leakages which should be avoided. As such, the transparency and provenance of the learning process should be delivered without revealing individual information, which may be obtained by adopting FL-native controls such as XAI local to the nodes~\cite{herrera2025reflections} and information signed via pseudonymous via the usage of ZKP techniques.~\cite{li2025integrating}} \\
\rewrite{Stakeholders should agree on the disclosure levels of information via assets such as model cards where thresholds are stated. This model cards could be later joined by accountable trails formed by the aforementioned information signed via pseudonymous as an evidence of the compliance to the agreement and its fulfillment.}

\paragraph{\textbf{Fairness vs. Privacy}}
\rewrite{While minority group information is key for proper generalization and achieving fairness, it could lead to potential attack vectors where information about these groups. This is a well known problem in FL. In this situation, again, calibration of privacy preserving FL-native controls like DP joined by the application of methods (e.g.: the ones mentioned in Section~\ref{sec3:challenges}) for heterogeneous data is crucial~\cite{noble2022differentially}.} \\
\rewrite{Here, stakeholders should agree on the privacy budget used and which techniques to use. Like in previous cases, red teaming experiments could help to provide an extensive analysis that would help to find proper parameters. This could again be used as an evidence for proper accountability. Also, in order to not disclosure sensitive attributes in the generated evidence, this should only contain DP protected aggregates, enabling insights without exposure.}

\paragraph{\textbf{Accountability vs. Privacy}}
\rewrite{So far in this section, we have shown how to provide evidence that preserves privacy and fulfill others TAI pillars while encourages accountability. However, as already mentioned, this evidence may lead to privacy leakage vectors if not carefully crafted and planned. It is key to take this into account when producing any kind of trail about the federated process. Thus, one should be selective about what metadata to disclose~\cite{mugunthan2020blockflow}.} \\
\rewrite{We suggest that stakeholders should define who can see which information, for how long and also provide reasons why up front. This shall be reflected in Data Protection Impact Assessment documents explaining the reached agreement. In addition, the inclusion third party audit assessments would increase the Trust perceived by the stakeholders and thus of Trust of the whole system.}

\paragraph{\textbf{Trustworthiness as a Unified System}}
\major{Viewed through TAI, \rewrite{these requirements function as indispensable pillars, and so the failure of one undermines the integrity of the whole system}. The seven dimensions must be satisfied simultaneously in order to achieve a trustworthy system. Improvements in one dimension \rewrite{determines} the feasible set \rewrite{of improvements} in others, \rewrite{leading to the need of carefully defined design choices}. \rewrite{Thus,} developing TFL requires an integrated approach that aligns FL-native mechanisms, \rewrite{such as the ones described above}, with governance and HITL processes so that trustworthiness, as a system-level property, \rewrite{is achieved}.} \\

\ph{As illustrated in Table~\ref{tab:Blueprint_Decisions}, decisions such as objective changes or automated retraining inherently involve trade-offs across robustness, fairness, and accountability, and therefore require explicit governance and evidence beyond local optimization criteria.}

\paragraph{\textbf{\ph{From coordination logic to auditable evidence}}}
\ph{The coordination blue\-print defines how trust-relevant decisions are made, justified, and governed in FL systems. The Trust Report~2.0 introduced in the next section complements this blueprint by defining the evidence demonstrating that such decisions were made in accordance with the requirements of TAI. In this sense, the blueprint specifies the decision logic and governance structure, while the Trust Report~2.0 provides the observable, auditable evidence surface through which trustworthiness can be continuously assessed. This is reflected in Figure \ref{fig:Blue_Trust}}.

\begin{figure}[th!]
\centering
\includegraphics[width=0.6\linewidth]{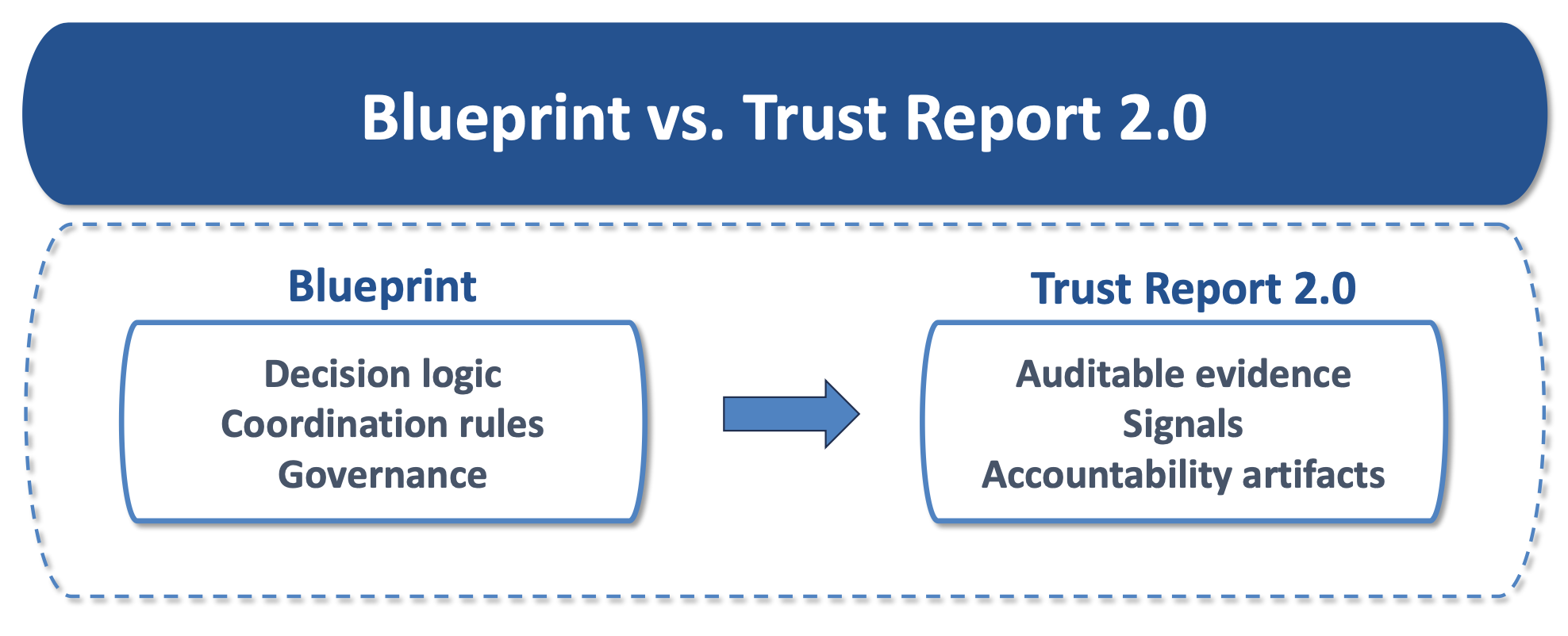}
\caption{\ph{Coordination logic to auditable evidence.}}
\label{fig:Blue_Trust}
\end{figure}

\ph{The evidence artifacts associated with each decision type in Table~\ref{tab:Blueprint_Decisions} directly inform the structure of the Trust Report~2.0, which operationalizes these coordination requirements as auditable and continuously updated signals.}

\subsection{\ph{Trust Report 2.0: Decision-Centric Evidence for Trustworthy Federated Learning}}

Trustworthiness in FL depends not only on technical guaranties, but also on the ability of stakeholders to understand, assess, and justify system behavior over time. In agentic and dynamic settings, where key lifecycle decisions can be taken autonomously, this requires structured and privacy-preserving evidence that explains what decisions were made, why they were made, and under which constraints. To this end, we propose a lightweight Trust Report~2.0 as the shared governance artifact that complements the coordination blueprint by surfacing auditable trust signals without centralizing raw data.

The Trust Report~2.0 is designed to (i) expose the state of cross-requirement trade-offs in a comparable and interpretable form, (ii) enable early warning and course correction through continuous monitoring, and (iii) support accountability by binding TAI requirements to FL-native controls and minimal anonymized provenance. Importantly, the Trust Report should not be understood as an additional compliance burden, but as a coordination device: it aligns acceptance criteria, thresholds, and release policies among clients, coordinators, and regulators, while ensuring that only privacy-preserving information leaves local nodes.

We propose a template for a minimum Trust report that contains the following points:
\begin{itemize}
    \item {\minor{Actor:}} The Trust report should contain \rewrite{the information agreed by the stakeholders.} \minor{We distinguish (i) client-produced Local Trust Packets (LTPs) containing privacy-thrifty, signed evidence per round; (ii) a coordinator-produced Trust Report compiled from LTPs and global aggregation operational metrics; and (iii) optional third-party assurance (e.g., an audit appendix) that verifies selected claims without accessing raw records.}
    \item {Metrics:} \minor{We distinguish between mandatory and optional Trust Report metrics to balance comparability across deployments with domain flexibility. Mandatory metrics include: (i) run metadata and participation, (ii) privacy accounting (including $\epsilon, \delta$ and cumulative budget when DP is used), (iii) robustness/security operational indicators (enabled defenses; aggregated flagging/rejection rates), (iv) utility/performance with an acceptance or non-regression criterion, and (v) system cost proxies (communication overhead and round latency). Optional metrics cover domain-specific or constraint-dependent signals (e.g., fairness or interpretability); if omitted, the Trust Report should briefly state the reason (eg., unavailable under privacy/legal constraints).}
    \item {Cadence:} \minor{Cadence is dual, a per-round report supports operational monitoring, while a per-release report consolidates evidence for governance and external accountability (with optional weekly digests for non-technical stakeholders).}
\end{itemize}

\ph{Together, the coordination blueprint defines the decision logic of TFL systems, while the Trust Report~2.0 provides the observable evidence surface through which those decisions can be assessed and governed. The practical instantiation of this Trust Report is illustrated in Section \ref{sec:TAI-Health} through a multi-hospital FL scenario.}

\ph{Similarly, proposals for tamper-evident coordination metadata (e.g., block-chain backed audit trails) illustrate why decision provenance should be treated as first-class evidence, separate from model-centric logs~\cite{Moore2025}.}

\paragraph{\ph{Decision-to-evidence alignment}}
\ph{To ensure that the Trust Report~2.0 operationalizes the coordination blueprint rather than duplicating it, each class of trust-relevant decision identified in Table~\ref{tab:Blueprint_Decisions} is explicitly associated with a minimal set of evidence fields. This alignment clarifies how governance decisions translate into auditable signals, and ensures that evidence collection remains proportional, privacy-preserving, and directly tied to TAI requirements.}

\ph{Table~\ref{tab:Decision_TR_Alignment} makes this alignment explicit by mapping each coordination decision to the concrete evidence fields that must be surfaced in the Trust Report~2.0. Rather than prescribing detailed logging schemas or implementation-specific mechanisms, the table defines a minimal and decision-centric evidence surface that is sufficient to support accountability, oversight, and explainability across deployments. By constraining evidence collection to information that is both privacy-preserving and directly motivated by governance needs, this mapping ensures that Trust Report~2.0 remains lightweight while still enabling meaningful assessment of trustworthiness over time.}

\begin{table}[ht!]
\centering
\begin{colortabular}{black}
\begin{tabular}{p{3.4cm} p{7.2cm} p{2.0cm}}
\toprule
\textbf{Decision Type} &
\textbf{Trust Report~2.0 Evidence Fields} &
\textbf{Primary TAI Req.} \\
\midrule
Client admission or exclusion &
Decision rationale, applicable policy identifier, cohort-level participation statistics, timestamped approval or escalation record &
R1, R3, R5 \\[0.3em]

Objective or metric change &
Change justification, prior and updated objective identifiers, expected impact summary, authorization record &
R2, R5, R7 \\[0.3em]

Model release or rollback &
Evaluation summary, non-regression checks, approval log, deployed model version identifier &
R1, R2, R7 \\[0.3em]

Automated retraining trigger &
Drift indicators, trigger thresholds, retraining event log, affected rounds &
R2, R5, R6 \\[0.3em]

Privacy budget update &
Privacy accounting snapshot ($\epsilon$, $\delta$, cumulative usage), risk assessment summary, authorization record &
R3, R7 \\
\bottomrule
\end{tabular}
\end{colortabular}
\caption{\ph{Alignment between coordination decisions (Table~\ref{tab:Blueprint_Decisions}) and Trust Report~2.0 evidence fields.}}
\label{tab:Decision_TR_Alignment}
\end{table}

\ph{Together, Tables~\ref{tab:Blueprint_Decisions} and~\ref{tab:Decision_TR_Alignment} illustrate how coordination decisions and evidence production are jointly structured, enabling trustworthiness to be assessed as a continuous, system-level property rather than as a collection of isolated guarantees.}

\paragraph{\ph{Operational use of the coordination blueprint}}
\ph{While the coordination blueprint and Trust Report~2.0 are presented at a conceptual level, they are intended to function as practical governance instruments across the federated lifecycle. At design time, system designers use the blueprint to define policies, thresholds, and autonomy boundaries, explicitly linking TAI requirements to permissible decision classes and required evidence. During training and operation, system operators rely on per-round Trust Report signals to monitor compliance, detect drift or anomalies, and trigger escalation or corrective actions without accessing raw data. At release time, governance bodies and regulators review consolidated Trust Reports to assess whether acceptance criteria are met, trade-offs are justified, and responsibilities are clearly assigned. In this way, the blueprint structures who decides what, when, and based on which evidence, ensuring that trustworthiness is not merely specified in principle but enacted through repeatable and auditable practice.}

\ph{This work provides an organizing and operationalization framework for TFL; empirical evaluation of specific instantiations of the proposed blueprint and Trust Report~2.0 is an important direction for future work.}

\section{\ph{Healthcare as a Stress-Test Domain for Trustworthy Federated Learning}}

\label{sec:TAI-Health}

\ph{Healthcare provides a uniquely demanding environment to examine whether TFL can move beyond abstract principles and function as a governable system in practice. Clinical deployments combine highly sensitive data, strong regulatory oversight, dynamic operating conditions, and heterogeneous institutional responsibilities, making them particularly susceptible to failures of coordination, accountability, and trust. Rather than serving as an illustrative use case, healthcare is therefore used in this section as a stress test: a domain in which the limitations of privacy-centric guarantees become visible, and where the necessity of lifecycle evidence, explicit governance, and multi-stakeholder alignment can be rigorously assessed.}

\ph{The following subsections examine this stress test at two complementary levels. First, we consider healthcare FL from a systemic perspective, highlighting how regulatory pressure, organizational plurality, and environmental dynamics challenge traditional notions of trust. Second, we focus on oncology FL as an extreme scenario that concentrates these pressures, and use it to validate how the proposed coordination blueprint and Trust Report~2.0 can operationalize trustworthiness under real clinical and governance constraints.}

\ph{We emphasize that the healthcare and oncology deployments discussed in this section are not intended as experimental validations of specific algorithms, but as stress-test scenarios for trust governance. Their role is to assess whether the proposed coordination blueprint and Trust Report~2.0 surface the right categories of decision-centric evidence under stringent regulatory, clinical, and organizational constraints. In this sense, Trust Report fields are validated by plausibility, governance alignment, and completeness of accountability coverage, rather than by comparative performance benchmarking.}

\subsection{\ph{Trustworthy FL under Clinical, Regulatory, and Organizational Constraints}}
\label{subsec:TFL_Healthcare_System}

\ph{Healthcare represents one of the most demanding deployment contexts for FL, combining highly sensitive data, strict regulatory oversight, and multi-stakeholder governance across institutional boundaries. Recent surveys highlight that, while FL is widely adopted in digital health to address privacy constraints, its practical deployment raises broader trust challenges related to accountability, transparency, and governance that extend beyond data protection alone~\cite{kanauzia2026comprehensive,eden2025scoping}. In this sense, healthcare provides a natural stress test for TFL, as failures in coordination, oversight, or evidence can have direct clinical, legal, and ethical consequences.}

\ph{From a technical perspective, healthcare FL systems operate in intrinsically dynamic environments. Clinical data distributions evolve over time due to changes in patient populations, diagnostic protocols, medical devices, and treatment guidelines, introducing temporal drift that directly affects model validity and safety. At the same time, regulatory frameworks such as GDPR and HIPAA impose ongoing obligations regarding consent, purpose limitation, and auditability, which cannot be discharged through static, design-time guarantees. As noted in recent governance reviews, current FL deployments often lack standardized mechanisms to document how trust-relevant decisions are made and justified over time, particularly when adaptive or semi-autonomous system behavior is introduced~\cite{eden2025scoping}.}

\ph{These system-level proposals move beyond abstract discussions of trust by illustrating how security mechanisms, governance structures, and operational coordination must be jointly realized in concrete healthcare FL deployments. In Medical IoT settings, blockchain-FL combinations have been proposed to support trustworthy medical data sharing, reinforcing the convergence of security, governance, and interoperability constraints in healthcare-grade federated deployments~\cite{gan2026dualdriven}. Operational FL deployments in European clinical networks have been proposed as platform-level services (e.g., for stroke management), illustrating the practical need for multi-center governance, secure coordination, and auditable lifecycle decisions under real clinical constraints~\cite{santos2025stroke}.}

\ph{Crucially, healthcare FL is inherently multi-stakeholder: hospitals, clinicians, patients, technical operators, and regulators all participate in or are affected by system decisions. This plurality of roles amplifies the need for shared governance artifacts that make trade-offs between privacy, utility, robustness, and fairness visible and contestable. As argued in the healthcare FL literature, trust in such systems depends not only on cryptographic or statistical guarantees, but also on the availability of interpretable, auditable evidence that supports clinical accountability and institutional responsibility~\cite{kanauzia2026comprehensive}. These characteristics make healthcare an appropriate domain to examine whether the coordination blueprint and Trust Report~2.0 proposed in this paper are sufficient to operationalize trustworthiness in practice.}

\ph{Recent clinical perspectives on agentic AI emphasize the role of orchestration, AI literacy, and human oversight in dynamic, high-risk medical settings, underscoring that trust in healthcare AI depends as much on governance and decision transparency as on technical performance~\cite{hanna2026infectious}.}

\subsection{\ph{Oncology Federated Learning as a Trust Stress Scenario}}
\label{subsec:Oncology_Stress}

\ph{Within healthcare, oncology represents a particularly stringent stress scenario for TFL due to the combination of high clinical risk, complex data modalities, and strong privacy and governance constraints. Recent surveys emphasize that effective AI-driven oncology requires access to large, diverse, multi-institutional datasets spanning imaging, genomics, clinical records, and treatment outcomes, while simultaneously complying with strict privacy regulations and institutional data silos~\cite{qi2026federated}. FL has therefore emerged as a key enabling paradigm for collaborative oncology research, yet its deployment exposes the full spectrum of trust challenges discussed in this paper.}

\ph{Oncological FL systems are characterized by pronounced data heterogeneity, longitudinal drift, and evolving clinical objectives. Diagnostic criteria, treatment protocols, and outcome definitions may change over time, requiring adaptive model updates and coordination decisions that directly affect patient care. As highlighted in oncology-specific studies, these dynamics exacerbate trade-offs between robustness, fairness, and clinical validity, particularly when models are retrained or personalized across institutions with uneven resources and patient populations~\cite{qi2026federated}. In such settings, the absence of explicit governance mechanisms and auditable decision traces can undermine clinical confidence even when model performance appears satisfactory.}

\ph{Recent work on trustworthy and explainable FL in oncology further underscores the need for system-level evidence that links model behavior to clinical and organizational accountability. For example, hierarchical and explainable FL frameworks for leukemia diagnosis demonstrate that interpretability and transparency are essential for clinical acceptance, but are insufficient without complementary governance artifacts that document training conditions, update decisions, and evaluation boundaries~\cite{pervez2025towards}. Similarly, incentive and reward mechanisms for trustworthy medical FL highlight that trust emerges from aligned objectives and documented compliance, not from performance metrics alone~\cite{pandl2025reward}.}

\ph{From a governance perspective, oncology FL amplifies regulatory pressure and accountability demands. Decisions regarding model release, rollback, or retraining may have direct implications for diagnosis or treatment recommendations, making it essential to justify such actions with clear, privacy-preserving evidence. The Trust Report~2.0 proposed in this paper can be instantiated in this context to surface decision-centric evidence, such as evaluation summaries, drift indicators, and authorization records, that supports both clinical oversight and regulatory review.}

\paragraph{\ph{Oncology as a stress test for TAI requirements}}
\ph{Rather than serving as an illustrative deployment, oncology FL can be used to stress-test the TAI requirements and the proposed Trust Report~2.0 by exposing how clinical risk, data heterogeneity, and regulatory constraints jointly pressure governance, coordination, and accountability mechanisms. Table~\ref{tab:Oncology_TAI_Stress} summarizes how oncology-specific stressors map to TAI requirements, the corresponding challenge categories identified in Section~\ref{sec3:challenges}, and the Trust Report evidence needed to sustain trust over time.}

\begin{table}[ht!]
\centering
\begin{colortabular}{black}
\begin{tabular}{p{3cm} p{2.5cm} p{2.5cm} p{4.2cm}}
\toprule
\textbf{Oncology Stressor} &
\textbf{Primary TAI Req.} &
\textbf{Relevant Challenges} &
\textbf{Trust Report~2.0 Evidence} \\
\midrule
High clinical risk and safety impact &
R2 Robustness \& Safety &
Ch.~2.1-2.3 &
Evaluation summaries, non-regression checks, safety thresholds, approval records \\[0.3em]

Multi-modal, heterogeneous data (imaging, genomics, EHR) &
R5 Fairness \& Non-discrimination &
Ch.~5.1-5.2 &
Subgroup performance indicators, heterogeneity notes, mitigation actions \\[0.3em]

Longitudinal drift in protocols and outcomes &
R2, R5 &
Ch.~2.X (drift), Ch.~5.X &
Drift indicators, retraining triggers, temporal validation summaries \\[0.3em]

Strict privacy and institutional silos &
R3 Privacy \& Data Governance &
Ch.~3.1-3.2 &
Privacy accounting ($\epsilon,\delta$), governance actions, unlearning records \\[0.3em]

Adaptive coordination and model updates &
R1 Human Oversight &
Ch.~1.X (autonomy boundaries) &
Decision rationales, autonomy level, escalation/approval logs \\[0.3em]

Regulatory accountability and audit pressure &
R7 Accountability &
Ch.~7.1-7.X &
Versioned Trust Reports, provenance claims, audit trail, role attribution \\
\bottomrule
\end{tabular}
\end{colortabular}
\caption{\ph{Oncology FL as a stress test for TAI requirements and Trust Report~2.0.}}
\label{tab:Oncology_TAI_Stress}
\end{table}

\ph{In this way, oncology FL serves not as a single use case, but as a comprehensive stress test that validates the necessity of treating trustworthiness as a continuous, governable system property.}

\section{\ph{From Privacy to Trust in Trustworthy Federated Learning}}
\label{Sec:TAI-Privacty-to-Trust}

\ph{Recent syntheses of Trust in AI underline that trust is not reducible to technical performance, but is instead relational, risk-laden, and grounded in situations of vulnerability and dependence. In particular, trust and trustworthiness can diverge in practice: a system can satisfy formal criteria while failing to inspire justified reliance among its stakeholders~\cite{afroogh2024trust,HENRIQUE2024100043}. These analyses situate trust within broader normative expectations and motivate a re-frame of TFL beyond privacy preservation alone.}

\ph{In this paper, we distinguish between \emph{trustworthiness} as an evidence-backed, system-level operating condition, and \emph{trust} as the justified reliance of stakeholders on that condition, shaped by context, risk, and governance expectations.}

\ph{Following \cite{duran2025trust}, we adopt a conception of trust in AI as a relational stance that combines justified reliance with normative expectations such as responsibility, responsiveness, and accountability. Applied to FL, this view implies that trustworthiness cannot be exhausted by privacy guarantees, robustness claims, or accuracy metrics. Instead, it must also be legible in how roles, duties, and recourse are distributed across clients, coordinators, and oversight actors. Trust in TFL thus emerges as a system-level condition when computational safeguards (e.g., privacy accounting, robust aggregation, red-teaming) are coupled with governance mechanisms (e.g., role-segmented oversight, escalation paths, and auditable decision records) that align with stakeholders’ moral and institutional expectations.}

\ph{Viewed through this lens, trust in FP emerges not from isolated guarantees, but from lifecycle governance that binds requirements, controls, and evidence across both learning and control planes.}

\ph{In doing so, we connect the paper’s initial governance problem (who is responsible, under what evidence), operational problem (how assurance persists over time), and paradigm-shift problem (agentic AI, dynamic FL) to the concrete challenge structure and coordination artifacts developed in Sections 3–6.}

\ph{At the same time, recent work argues that TAI must be understood as context-sensitive and perceptual: compliance with abstract principles is necessary but insufficient if the situated experience of users does not support trust~\cite{wirz2025re}. This insight is particularly salient for FL, where risks, data distributions, and regulatory constraints differ across nodes. Reframing trust in this setting therefore requires making guarantees usable at the edge: node-local explanations that do not re-identify individuals; privacy budgets that can be interpreted and contested by local stakeholders; provenance claims that demonstrate policy compliance without exposing sensitive records; and federated fairness assessments that respect local sensitivities while enabling collective assurance. In this sense, TFL materializes when clinicians, regulators, engineers, and community representatives can each access the right evidence, at the right granularity, within their own operational context~\cite{wirz2025re}.}

\ph{These considerations reinforce a broader caution articulated in the TAI literature: policy aspirations such as lawfulness, ethical alignment, and robustness risk dilution if they are not operationalized through concrete, verifiable practices~\cite{stix2022artificial}. This paper’s position on TFL responds directly to that concern. Rather than treating trust as a label or a static property of trained models, we frame it as an operating condition sustained through repeatable processes and auditable artifacts. The coordination blueprint and Trust Report~2.0 introduced in this work translate high-level requirements into decision-centric governance structures and lifecycle evidence, enabling trust to be continuously assessed, negotiated, and renewed.}

\paragraph{\ph{The present: evidence-backed trust at the node}}
\ph{In current TFL deployments, trust takes the form of practical confidence that the system will behave as promised under real-world constraints. This confidence is earned when privacy, reliability, and responsibility are made legible where work actually occurs: at the federated node. Privacy is treated as a system property grounded in minimization, purpose limitation, and consent, while exposing just enough evidence to support justified reliance. Crucially, technical safeguards alone are insufficient. Clear roles, duties, and recourse mechanisms are required so that clinicians, regulators, engineers, and community stakeholders can exercise situated oversight without extracting raw data. When designed explicitly, tensions such as fairness versus efficiency, privacy versus robustness, or accountability versus confidentiality become negotiated design choices coordinated through shared thresholds, acceptance criteria, and release policies. Trust, in this sense, is not perfection, but disciplined constraint coupled with accountable proof.}

\paragraph{\ph{The future: trust as a continuously earned condition}}
\ph{Looking forward, trust in TFL must mature from episodic assurance into a continuously evidenced operating condition. This entails learning under limits: federated ecosystems in which trust is co-produced by technical controls and human governance, and calibrated to context rather than asserted by compliance alone. Such a future favors adaptive privacy budgets aligned with purpose and risk, context-aware consent at the edge, explanations that remain useful without re-identification, and fairness audits that surface population-level risks while keeping sensitive attributes local. It also requires accountability mechanisms that verify behavior without drifting into surveillance. Common metrics beyond accuracy, such as information leakage, robustness under attack, equity under differential privacy noise, transparency coverage, provenance completeness, incident response latency, and sustainability overhead, must be reported through recurring Trust Reports. As these practices normalize, privacy becomes the enabler of collaboration rather than its obstacle: institutions can co-create models across organizational and national boundaries because the limits they refuse to cross are explicit, enforced, and continuously evidenced. In this way, trust in FL becomes not an aspiration, but the everyday discipline through which collaboration remains possible.}

\section{\ph{Conclusions: Toward Trustworthy Federated Learning in the Agentic Era}}
\label{sec:final}

\ph{This paper set out to re-examine FL through the lens of trust rather than privacy alone. While privacy preservation remains a necessary foundation, our analysis shows that it is no longer sufficient in the presence of agentic behavior, dynamic environments, and multi-stakeholder governance. In such settings, trustworthiness cannot be reduced to isolated technical guarantees, but must instead be understood as a system-level, lifecycle-dependent operating condition.}

\ph{To support this reframing, we introduced a requirement-driven taxonomy of challenges grounded in TAI principles and extended to account for autonomy, control-plane decision-making, and environmental non-stationarity. This taxonomy clarifies not only individual trust challenges, but also the tensions and trade-offs that arise when multiple requirements interact under decentralization and adaptation. Building on this diagnosis, we proposed a coordination blueprint that focuses on how trust-relevant decisions are governed, justified, and aligned across stakeholders, rather than on prescribing specific learning mechanisms.}

\ph{A central contribution of this work is the Trust Report~2.0, which operationalizes trustworthiness as auditable, decision-centric evidence produced throughout the federated lifecycle. By separating coordination logic from evidence surfacing, the Trust Report enables accountability and oversight without undermining privacy or decentralization. The healthcare domain, and oncology FL in particular, served as a stress test that demonstrates why such lifecycle evidence and governance structures are indispensable in practice.}

\ph{Ultimately, this work shows that sustaining trust in FL depends on decision-centric governance that aligns requirements, federated controls, and auditable evidence under explicit lifecycle constraints.
 More broadly, our results suggest that the future of FL lies not in ever-stronger pointwise guarantees, but in disciplined coordination under explicit limits. Treating trust as an operating condition shifts the focus from certifying models to governing systems, from static assurances to recurring validation, and from abstract principles to verifiable practice. We believe this perspective provides a necessary foundation for deploying FL responsibly in the agentic era and opens the door to future work on standardized evidence interfaces, interoperable governance artifacts, and domain-specific trust calibration.}

 \paragraph{\ph{Limitations and scope}} \ph{This work deliberately prioritizes conceptual clarity, governance structure, and operationalization over algorithmic novelty or empirical benchmarking. As such, the proposed taxonomy, coordination blueprint, and Trust Report~2.0 do not prescribe specific technical mechanisms, nor do they evaluate particular instantiations against quantitative baselines. This choice reflects the paper’s scope: to provide reusable conceptual infrastructure that can accommodate diverse FL architectures, threat models, and regulatory contexts. Future work may instantiate and empirically evaluate these structures in specific domains, but the primary contribution here lies in making trust a governable, decision-centric system property rather than an implicit byproduct of isolated technical guarantees.}

\ph{Taken together, this work provides a unifying perspective on FL in the agentic era by reframing trustworthiness as a lifecycle-dependent, system-level operating condition. By distinguishing between learning and control planes and centering governance on decisions rather than isolated model properties, the paper introduces an organizing abstraction that aligns TAI requirements with federated controls and auditable evidence. This perspective translates trust from an abstract aspiration into concrete structures (requirements, decision processes, evidence artifacts, and governance roles) without compromising decentralization or privacy. While healthcare and oncology serve as a stress-test domain, the proposed taxonomy, coordination blueprint, and Trust Report~2.0 are intentionally domain-agnostic and applicable to a wide range of decentralized, multi-stakeholder settings. Ultimately, sustaining trust in FL depends not on ever-stronger pointwise guarantees, but on disciplined, decision-centric governance under explicit lifecycle constraints, shifting the focus from certifying models to governing systems through continuous, verifiable practice.
}

\ph{Viewed through the lens of Trust Report 2.0, trust in FL emerges not as a static guarantee derived from privacy mechanisms, but as a continuously evidenced property shaped by system decisions, governance constraints, and operational context.}

\section*{Acknowledgments}

\minor{This publication is part of the TSI-100927-2023-1 Project, funded by the Recovery, Transformation and Resilience Plan from the European Union Next Generation through the Ministry for Digital Transformation and the Civil Service.}

\clearpage
\bibliographystyle{unsrt}  
\bibliography{references}  

\clearpage

\appendix

\section{\ph{Positioning with Respect to Existing Surveys}}
\label{sec:related_work}

\ph{The literature on FL and its trust implications has expanded rapidly in recent years, resulting in a diverse set of surveys that approach the problem from different perspectives. Rather than providing an exhaustive narrative review, this section organizes representative surveys into conceptual blocks and analyzes them through the lens of trustworthiness. Specifically, we distinguish between: (i) technical surveys on FL mechanisms and optimization; (ii) requirement-oriented surveys that address TAI principles in FL; (iii) domain-specific surveys, with a focus on healthcare applications; and (iv) emerging surveys on agentic, LLM-enabled, and dynamic learning systems. }

\paragraph{\ph{Reading guide for Appendix A}}
\ph{To keep the survey comparison analytical rather than encyclopedic, we organized previous work using two orthogonal tags. First, each reference is assigned a \emph{Paper Type} indicating its role in the literature (e.g., technical FL survey, requirement-oriented TFL survey, healthcare governance review, or agentic/LLM/dynamic overview). Second, each reference is assigned a \emph{Primary Focus} tag, which acts as a semantic classifier capturing the dominant perspective of the work (i.e., what problem it treats as central). These tags are intentionally coarse-grained: they support fast clustering and gap identification, and they make explicit where prior surveys typically underemphasize agency, governance, and lifecycle evidence relative to the contributions developed in Sections~\ref{sec3:challenges}-\ref{Sec:TAI-Blueprint}. Table \ref{tab:survey-key-paper-type-focus} shows these coverage. }

\begin{longtable}{@{}p{0.2\linewidth} p{0.25\linewidth} p{0.5\linewidth}@{}}
\caption{\minor{\ph{Reading key for Section~\ref{sec2:preliminaries} survey tables.}}}
\label{tab:survey-key-paper-type-focus} \\

\toprule
\textbf{Paper Type} & \textbf{Primary Focus} & \textbf{Typical Coverage} \\
\midrule
\endfirsthead

\multicolumn{3}{c}{\tablename\ \thetable{} continued from previous page} \\
\toprule
\textbf{Paper Type} & \textbf{Primary Focus} & \textbf{Typical Coverage} \\
\midrule
\endhead

\midrule
\multicolumn{3}{r}{} \\
\endfoot

\bottomrule
\endlastfoot

\minor{Technical FL surveys}
& \minor{Learning-centric} 
& \minor{Core FL formulation, training pipelines, convergence-oriented framing.}\\

& \minor{Optimization-centric} 
& \minor{Algorithms, aggregation rules, optimization .} \\

& \minor{Privacy/Security-centric} 
& \minor{DP, secure aggregation, inference risks, adversarial models.} \\

& \minor{Systems-centric} 
& \minor{Scalability, communication,  deployment constraints.}\\

& \minor{Robustness-centric} 
& \minor{Reliability under attacks, failures, generalization under shift.} \\

& \minor{Heterogeneity-centric} 
& \minor{Non-IID analysis, personalization, client variability.} \\

& \minor{Fairness-centric} 
& \minor{Fairness analysis.} \\

\midrule

\minor{TFL / requirement oriented surveys}
& \minor{Requirement-centric} 
& \minor{Challenge framing aligned to TAI-like requirements (privacy, fairness, transparency, etc.).} \\

& \minor{Property-centric} 
& \minor{Trustworthiness reduced to selected properties (e.g., privacy, robustness).} \\

& \minor{Trust-dimension-centric} 
& \minor{Trust attributes treated as discrete dimensions with partial mappings to FL.}\\

& \minor{Challenge-taxonomy} 
& \minor{Taxonomies and roadmaps of challenges. }\\

\midrule

\minor{Healthcare / domain specific works}
& \minor{Domain-driven} 
& \minor{Clinical applications, privacy constraints, feasibility in healthcare workflows.}\\

& \minor{Regulation-driven} 
& \minor{Compliance framing, risk management.}\\

& \minor{Clinical-deployment-centric} 
& \minor{Deployment pathways, stakeholder roles, adoption barriers in clinical settings.}\\

& \minor{Governance-aware} 
& \minor{Accountability,  institutional arrangements.}\\

\midrule

\minor{Agentic / LLM / dynamic learning }
& \minor{Agency-centric} 
& \minor{Agent architectures, coordination, autonomous decision-making and planning.} \\

& \minor{Autonomy-centric} 
& \minor{Autonomy levels, delegation, policy boundaries, oversight interfaces.} \\

& \minor{Dynamic systems centric} 
& \minor{Temporal drift, continual learning, non-stationarity and adaptation strategies.} \\

& \minor{LLM orchestration centric} 
& \minor{Distributed LLM reasoning, coordination, semantic processing pipelines.}\\

& \minor{Control-centric} 
& \minor{Orchestration,  multi-agent workflow control.} \\

\end{longtable}

\paragraph{\ph{Scope and representativeness}}
\ph{Given the rapid growth of the FL and TAI literature, this section does not aim to provide an exhaustive catalog of all existing surveys and overviews. Instead, we select a set of representative and recent works that collectively reflect the dominant perspectives shaping the field. The surveys included below are chosen to illustrate recurring patterns, emphases, and blind spots across technical, requirement-oriented, domain-specific, and agentic or dynamic viewpoints.
For each block associated to ``paper type'', we summarize recent contributions in the respective table, and highlight which dimensions of trust are addressed or overlooked. 
As such, the tables should be read as an analytical lens for positioning this work, rather than as a comprehensive bibliography of prior surveys.They must be  used to clarify how existing surveys collectively advance the field, and to motivate the need for the requirement-driven, agency-aware, and decision-centric perspective adopted in this work:}

\paragraph{\ph{Technical surveys on FL mechanisms and optimization}}
 \ph{Table  \ref{tab:technical-fl-surveys} groups representative works that primarily model FL as a \emph{learning pipeline}, emphasizing optimization strategies, robustness, system efficiency, and communication constraints. These surveys and conceptual studies provide essential foundations for scalable FL deployment, but they typically treat trust implicitly, often equating it with privacy preservation or robustness against specific attacks, while offering limited treatment of governance structures, agency-aware decision-making, or lifecycle-oriented trust evidence. As such, they motivate but do not resolve the broader shift toward TFL developed in subsequent sections.}

\begin{longtable}{@{}p{0.15\linewidth} p{0.10\linewidth} p{0.45\linewidth} p{0.2\linewidth}@{}}
\caption{\minor{Representative technical surveys on FL.}}
\label{tab:technical-fl-surveys} \\

\toprule
\textbf{Survey} & \textbf{Year} & \textbf{Description} & \textbf{Primary Focus} \\
\midrule
\endfirsthead

\multicolumn{4}{c}{\tablename\ \thetable{} continued from previous page} \\
\toprule
\textbf{Survey} & \textbf{Year} & \textbf{Description} & \textbf{Primary Focus} \\
\midrule
\endhead

\midrule
\multicolumn{4}{r}{} \\
\endfoot

\bottomrule
\endlastfoot

Salazar et al.~\cite{salazar2026survey} & 2026 & \minor{Survey on group fairness in FL, providing a taxonomy of fairness challenges, solution approaches, and future research directions.} & \minor{Fairness-centric} \\

Ji et al.~\cite{ji2024emerging} & 2024 &
\minor{Discussion of emerging challenges and future research directions for FL motivated by real-world deployment constraints and application scenarios.} &
\minor{Heterogeneity-centric} \\

Wen et al.~\cite{wen2023survey} & 2023 &
\minor{Comprehensive overview of FL fundamentals, including optimization algorithms, privacy and security mechanisms, communication efficiency, and data heterogeneity.} &
\minor{Learning-centric} \\

Qin et al.~\cite{qin2025knowledge} & 2025 &
\minor{Focused survey on knowledge distillation techniques in FL, addressing communication efficiency and model compression challenges. The scope is intentionally narrow and centered on a specific optimization mechanism rather than end-to-end trustworthiness or governance.} &
\minor{Optimization-centric} \\

Huang et al.~\cite{huang2024federated} & 2024 &
\minor{Survey of FL methods addressing generalization, robustness, and fairness under heterogeneous and non-IID data distributions.} &
\minor{Robustness-centric} \\

Uddin et al.~\cite{uddin2025systematic} & 2025 &
\minor{Systematic literature review of robust FL, analyzing issues, solution strategies, and future research directions.} &
\minor{Robustness-centric} \\

Chen et al.~\cite{chen2025advances} & 2025 &
\minor{Survey of robust FL methods with explicit consideration of data and system heterogeneity, covering robustness techniques and open problems.} &
\minor{Robustness-centric} \\

Feng et al.~\cite{feng2025survey} & 2025 &
\minor{Survey of security threats in FL, analyzing attack models, vulnerabilities, and mitigation strategies across different system assumptions.} &
\minor{Security-centric} \\

Li et al.~\cite{li2025threats} & 2025 &
\minor{Comprehensive survey of threats and defenses across the FL lifecycle, covering attack surfaces, protection mechanisms, and open challenges from training to deployment.} &
\minor{Security-centric} \\

Jimenez-Gutierrez et al.~\cite{jimenez2026security} & 2026 &
\minor{Extensive survey of security and privacy in FL, including attacks, defenses, frameworks, applications, and future research directions.} &
\minor{Privacy/Security-centric} \\

Nasim et al.~\cite{nasim2025principles} & 2025 &
\minor{Survey of FL architectures, outlining core components, system design principles, and architectural variations.} &
\minor{Systems-centric} \\

Ye et al.~\cite{ye2025vertical} & 2025 &
\minor{Survey of vertical FL methods, focusing on effectiveness, security, and practical applicability.} &
\minor{Systems-centric} \\

Wu et al.~\cite{wu2025vertical} & 2025 &
\minor{Practical survey of vertical FL, discussing real-world challenges, limitations, and deployment lessons.} &
\minor{Systems-centric} \\

Li et al.~\cite{li2025human} & 2025 &
\minor{Conceptual and application-oriented study introducing human-machine hybrid FL, where human inputs and automated optimization are jointly leveraged during training and deployment. While relevant to agency-aware FL, the work does not provide a requirement-driven or governance-oriented trust framework.} &
\minor{Systems-centric} \\

Quan et al.~\cite{quan2025federated} & 2026 &
\minor{Comprehensive survey of FL for cyber-physical systems, addressing architectural, communication, and deployment challenges.} &
\minor{Systems-centric} \\

\end{longtable}

\paragraph{\ph{TFL and requirement-oriented surveys}}
\ph{More recent surveys explicitly adopt a TAI perspective, organizing challenges in FL around dimensions such as privacy, robustness, fairness, transparency, and accountability. While these works represent an important step toward principled trust analysis, they often remain centered on requirement lists or static guarantees. In particular, agency, control-plane decision-making, and the operationalization of trust through continuous, auditable evidence are rarely addressed in a systematic way. Table \ref{tab:tfl-requirement-surveys} compares representative requirement-oriented surveys with respect to these missing primitives.}

\begin{longtable}{@{}p{0.15\linewidth} p{0.10\linewidth} p{0.45\linewidth} p{0.2\linewidth}@{}}
\caption{\minor{Representative surveys adopting a trustworthy or requirement-oriented perspective on FL. While these works organize challenges around trust-related dimensions, they typically provide limited treatment of agency, control-plane decision-making, and continuous lifecycle evidence.}}
\label{tab:tfl-requirement-surveys} \\

\toprule
\textbf{Survey} & \textbf{Year} & \textbf{Description} & \textbf{Primary Focus} \\
\midrule
\endfirsthead

\multicolumn{4}{c}{\tablename\ \thetable{} continued from previous page} \\
\toprule
\textbf{Survey} & \textbf{Year} & \textbf{Description} & \textbf{Primary Focus} \\
\midrule
\endhead

\midrule
\multicolumn{4}{r}{} \\
\endfoot

\bottomrule
\endlastfoot

Zhang et al.~\cite{zhang2023survey} & 2023 &
\minor{Survey of challenges in trustworthy FL, with emphasis on privacy, robustness, and security threats in distributed training.} &
\minor{Property-centric} \\

Tariq et al.~\cite{tariq2024trustworthy} & 2024 &
\minor{Requirement-oriented taxonomy of challenges in trustworthy FL, covering interpretability, transparency, privacy, robustness, fairness, and accountability.} &
\minor{Requirement-centric} \\

Zhang et al.~\cite{zhang2024survey} & 2024 &
\minor{Roadmap-style survey outlining research directions for trustworthy FL across explainability, fairness, privacy, and robustness.} &
\minor{Challenge-taxonomy} \\

Chen et al.~\cite{chen2025trustworthy} & 2025 &
\minor{Analysis of trustworthy FL with a primary focus on privacy and security mechanisms and their associated threats.} &
\minor{Privacy/Security-centric} \\

Jung \cite{jung2026trustworthyfl} & 2026 &
\minor{Tutorial-style overview of TFL, synthesizing key trust dimensions (privacy, robustness, fairness, accountability) and how they are typically operationalized.} &
\minor{Trust-dimension-centric)} \\

\end{longtable}

\paragraph{\ph{Agentic AI, LLM-enabled, and dynamic learning perspectives}}
\ph{A rapidly growing body of survey and vision literature addresses these scenarios. These works introduce concepts such as autonomous decision-making, orchestration, semantic communication, and continual adaptation, which fundamentally challenge the assumptions underlying classical FL. While most of these surveys are not framed explicitly within Trustworthy FL, they surface the core primitives, agency, autonomy, control, and dynamics, that motivate the agentic AI and lifecycle, aware extensions developed in Sections~\ref{sec3:challenges}-\ref{Sec:TAI-Blueprint} of this paper. We include surveys on foundation-model training in federated settings as they directly inform the feasibility and risk surface of LLM-enabled FL systems, even when they do not explicitly adopt an agentic AI framing.
Table \ref{tab:agentic-llm-dynamic-surveys}  positions representative overview works along these dimensions and clarifies their relevance to trustworthiness in the federated setting.}

\begin{longtable}{@{}p{0.15\linewidth} p{0.10\linewidth} p{0.5\linewidth} p{0.15\linewidth}@{}}
\caption{\minor{Representative overview works on agentic AI, LLM-enabled systems, and learning in dynamic environments. While not primarily focused on FL trustworthiness, these surveys surface key primitives (agency, autonomy, control, and dynamics) that motivate the extensions to to TFL developed in this paper.}}
\label{tab:agentic-llm-dynamic-surveys} \\

\toprule
\textbf{Overview} & \textbf{Year} & \textbf{Description} & \textbf{Primary Focus} \\
\midrule
\endfirsthead

\multicolumn{4}{c}{\tablename\ \thetable{} continued from previous page} \\
\toprule
\textbf{Overview} & \textbf{Year} & \textbf{Description} & \textbf{Primary Focus} \\
\midrule
\endhead

\midrule
\multicolumn{4}{r}{} \\
\endfoot

\bottomrule
\endlastfoot

Acharya et al.~\cite{acharya2025agentic} & 2025 &
\minor{Comprehensive survey of agentic AI systems, covering autonomous goal pursuit, planning, coordination, and decision-making architectures for complex tasks.} &
\minor{Agency-centric} \\

Giusti et al.~\cite{giusti2025federation} & 2025 &
\minor{Survey of semantics-aware communication fabrics for federations of agents, emphasizing coordination, interoperability, and distributed decision-making.} &
\minor{Agency-centric} \\

Nguyen et al.~\cite{nguyen2025vision} & 2025 &
\minor{Vision of agentic AI for FL, outlining workflows, system design, and emerging challenges under autonomy and dynamics.} &
\minor{Agency-centric} \\

Jiang et al.~\cite{jiang2025large} & 2025 &
\minor{Tutorial-style overview of the transition from large AI models to agentic AI, emphasizing autonomous reasoning, communication, and system-level coordination.} &
\minor{Autonomy-centric} \\

Sun et al.~\cite{sun2025vision} & 2025 &
\minor{Vision paper on decentralized agent swarm networks, focusing on coordination, autonomy, and productivity in large-scale.} &
\minor{Control-centric} \\

Li et al.~\cite{li2025position} & 2025 &
\minor{Position paper introducing agentic FL for strategy design, highlighting decision autonomy and coordination beyond parameter aggregation.} &
\minor{Control-centric} \\

Woisetsch-l{\"a}ger et al.~\cite{woisetschlager2024survey} & 2024 &
\minor{Survey of efficient FL methods for foundation model training, emphasizing communication/computation efficiency, scalability constraints, and practical system techniques for large models.} &
\minor{LLM-orchestration-centric} \\

Wei et al.~\cite{wei2025federated} & 2025 &
\minor{Survey of federated reasoning LLMs, analyzing distributed reasoning, coordination, and communication among LLM-based agents under decentralization.} &
\minor{LLM-orchestration-centric} \\

Kristiani et al.~\cite{kristiani2026deploying} & 2026 &
\minor{Survey of strategies for deploying LLM transformers on edge devices, including resource constraints, adaptation, and decentralized execution.} &
\minor{Deployment-centric} \\

Polato et al.~\cite{polato2026learning} & 2026 &
\minor{Tutorial on learning in federated and dynamic environments, covering concept drift, temporal non-stationarity, and adaptive strategies.} &
\minor{Dynamic-systems-centric} \\

Tang et al.~\cite{tang2026rethinking} & 2026 &
\minor{Survey of secure semantic communications in generative and agentic AI, addressing risks introduced by autonomy and semantic processing.} &
\minor{Security-centric} \\

Amini et al.~\cite{amini2025distributed} & 2025 &
\minor{Survey of distributed and multimodal LLMs, focusing on architectural choices, deployment challenges, and coordination across decentralized systems.} &
\minor{Systems-centric} \\

Sun et al.~\cite{sun2025generalizing} & 2025 &
\minor{Conceptual analysis of trust generalization in language models, examining how trustworthiness properties evolve under adaptation and deployment.} &
\minor{Trust-centric} \\

Sarker and Jesser~\cite{sarker2025advancing} & 2025 &
\minor{Overview of decentralized and verifiable collaboration mechanisms for agentic AI and foundation model development.} &
\minor{Governance-aware} \\

Raza et al.~\cite{raza2025trism} & 2025 &
\minor{Survey of trust, risk, and security management (TRiSM) for LLM-based agentic multi-agent systems.} &
\minor{Governance-aware} \\

\end{longtable}

\paragraph{\ph{Recent healthcare-focused perspectives}}
\ph{To complement the technical and re\-quire\-ment-oriented survey blocks in Appendix, we highlight a small set of recent healthcare-focused works (2025-2026) that are particularly relevant to TFL, see Table \ref{tab:healthcare-surveys}. These works emphasize clinical constraints, governance pressures, and incentive alignment, motivating the use of healthcare as a stress test domain for lifecycle trust evidence.  Rather than aiming for exhaustive coverage, this selection emphasizes how clinical deployment constraints, governance structures, and incentive alignment shape trust beyond privacy preservation. These studies help motivate why healthcare is treated in this paper as a stress test domain where multi-stakeholder accountability and lifecycle evidence become unavoidable. These works vividly illustrate why trust is indispensable in practice, yet they often treat trust implicitly, relying on domain norms rather than explicit governance and evidence mechanisms.}

\begin{longtable}{@{}p{0.15\linewidth} p{0.10\linewidth} p{0.5\linewidth} p{0.15\linewidth}@{}}
\caption{\minor{Selected recent healthcare-focused perspectives (2025-2026) relevant to TFL. These works emphasize clinical constraints, governance pressures, and incentive alignment, motivating the use of healthcare as a stress test domain for lifecycle trust evidence.}}
\label{tab:healthcare-surveys} \\

\toprule
\textbf{Survey /Study} & \textbf{Year} & \textbf{Description} & \textbf{Primary Focus} \\
\midrule
\endfirsthead

\multicolumn{4}{c}{\tablename\ \thetable{} continued from previous page} \\
\toprule
\textbf{Survey /Study} & \textbf{Year} & \textbf{Description} & \textbf{Primary Focus} \\
\midrule
\endhead

\midrule
\multicolumn{4}{r}{} \\
\endfoot

\bottomrule
\endlastfoot

Kanauzia et al.~\cite{kanauzia2026comprehensive} & 2026 &
\minor{Comprehensive survey of FL for privacy preservation in digital healthcare, covering application areas, privacy/security techniques, and deployment challenges under sensitive clinical data constraints.} &
\minor{Domain-driven} \\

Eden et al.~\cite{eden2025scoping} & 2025 &
\minor{Scoping review of governance in healthcare FL, focusing on organizational arrangements, accountability structures, and the practical challenges of deploying FL under regulatory and institutional constraints.} &
\minor{Governance-aware} \\

Pandl et al.~\cite{pandl2025reward} & 2025 &
\minor{Study of reward and incentive systems for trustworthy medical FL, analyzing how stakeholder incentives, participation dynamics, and compliance considerations shape trust and sustainability.} &
\minor{Governance-aware} \\

\end{longtable}

\ph{Beyond cataloging existing challenges, our analysis surfaces gaps that are often under-emphasized in prior surveys, particularly those related to human agency and oversight, societal and environmental considerations, and accountability, while still maintaining coverage of core technical issues.}

\paragraph{\ph{Positioning of this work}}
\ph{Taken together, the reviewed surveys provide a comprehensive account of advances and challenges in FL, yet they largely remain centered on isolated technical properties or static design-time guarantees. In particular, three foundational primitives required for sustaining trust in contemporary deployments remain underdeveloped.}

\begin{itemize}
\item \ph{First, existing works rarely adopt a requirement-driven perspective explicitly grounded in  TAI principles, limiting their ability to reason systematically about governance, accountability, and cross-requirement trade-offs.}

\item \ph{Second, prior surveys typically model FL  as a learning pipeline, overlooking the role of agency and control-plane decisions that arise in agentic, LLM-enabled, and dynamically adaptive systems. }

\item \ph{Third, while challenges are extensively cataloged, there is little guidance on how trustworthiness can be operationalized through decision-centric, auditable evidence produced across the federated lifecycle. }

\end{itemize}

\end{document}